%% file: main.tex
\documentclass{article}

\usepackage[main, final]{neurips_2026}

\usepackage[utf8]{inputenc} %
\usepackage[T1]{fontenc}    %
\usepackage{hyperref}       %
\usepackage{url}            %
\usepackage{booktabs}       %
\usepackage{amsfonts}       %
\usepackage{nicefrac}       %
\usepackage{microtype}      %
\usepackage{xcolor}         %
\newtheorem{definition}{Definition}[section]
\input{preamble}

\title{TAMEing the Open-World Personalization: Towards Open-Set Personalized MLLM Assistant}

\author{%
  Rongpei Hong$^{1}$ \quad
  Jian Lang$^{1}$ \quad
  Ting Zhong$^{1}$\thanks{Corresponding author: \texttt{zhongting@uestc.edu.cn}.}\phantom{$^{*}$} \quad
  Fan Zhou$^{1,2}$ \\
  $^{1}$University of Electronic Science and Technology of China \\
  $^{2}$Intelligent Digital Media Technology Key Laboratory of Sichuan Province \\
}

\begin{document}

\maketitle
\setcounter{footnote}{0} %

\input{sec/0-Abstract}

\input{sec/1-Introduction}

\input{sec/2-Related-Work}

\input{sec/3-Methodology}

\input{sec/4-Experiments}

\input{sec/5-Conclusion}

\begin{ack}
This work was supported by the National Natural Science Foundation of China (Grant No.~62572097).
This work was also supported by the Center for HPC, University of Electronic Science and Technology of China.
The authors declare no competing interests.
\end{ack}

\bibliographystyle{unsrt}
\bibliography{main}

\appendix

\input{sec/7-Appendix}
\stopcontents[appendix]

\end{document}

%% file: preamble.tex
\usepackage{newfloat}
\usepackage{listings}
\usepackage{titletoc}

\usepackage{algorithm}
\usepackage{algorithmic}
\usepackage{adjustbox}
\usepackage{color}

\usepackage{bm}
\usepackage{enumitem}
\usepackage[most]{tcolorbox}  %
\usepackage{amsmath,amsfonts}
\usepackage{float}

\usepackage{makecell}
\usepackage{todonotes}

\usepackage[table]{xcolor} %
\usepackage{colortbl}
\usepackage{arydshln}
\usepackage{xspace}
\usepackage{array}
\usepackage{multirow}
\usepackage{booktabs}
\usepackage{amssymb}
\usepackage{xcolor}
\usepackage{pifont}  %
\newcommand{\B}{LCMP\xspace}
\newcommand{\M}{TAME-O\xspace}

\newlength\savewidth
\newcommand\shline{\noalign{\global\savewidth\arrayrulewidth
                            \global\arrayrulewidth 1pt}%
                   \hline
                   \noalign{\global\arrayrulewidth\savewidth}}

\usepackage{cleveref}
\usepackage{color}
\usepackage{subcaption}
\definecolor{tablegreen}{RGB}{236, 255, 236}

\definecolor{occolor}{HTML}{f19d38} %
\newcommand{\oc}[1]{{\color{occolor}#1}} %

\usepackage{xcolor}
\definecolor{darkred}{rgb}{0.7,0,0}
\definecolor{darkgreen}{rgb}{0,0.5,0}
\usetikzlibrary{patterns}
\definecolor{tabred}{HTML}{D62728}
\definecolor{tabgreen}{HTML}{2CA02C}
\definecolor{tabgreenedge}{HTML}{1E7A1E}
\DeclareRobustCommand{\hatchbar}{\tikz[baseline=0ex]\filldraw[draw=tabred, pattern=north east lines, pattern color=tabred, line width=0.4pt] (0,0) rectangle (0.7em,0.7em);}
\DeclareRobustCommand{\solidbar}{\tikz[baseline=0ex]\filldraw[draw=tabgreenedge, fill=tabgreen, line width=0.4pt] (0,0) rectangle (0.7em,0.7em);}

%% file: sec/0-Abstract.tex
\begin{abstract}
Personalizing Multimodal Large Language Models (MLLMs) as intelligent assistants has recently emerged as an active research topic, where an MLLM is expected to go beyond generic, \textit{one-size-fits-all} replies, and instead generate responses grounded in user-specific objects, entities, and concepts encountered during interactions.
Nevertheless, existing studies mainly focus on personalizing MLLMs on a \textit{static, predefined set} of personalized concepts and require users to manually and frequently extend this set to accommodate newly emerging concepts, which is infeasible in dynamic-world scenarios.
To address this limitation, we recast conventional MLLM personalization as an \textbf{Open-Set problem}, and propose \textbf{\M}, the first long-context open-set personalized MLLM assistant.
Specifically, \M introduces a training-free, plug-and-play Concept Reconciliation Skill (\textbf{CR Skill}).
It enables the assistant to automatically unlock novel concepts during prolonged human–MLLM interactions while preventing confusion with previously unlocked ones, allowing the MLLM to continually grasp more and more concepts and progressively refine the modeling of known ones. 
Extensive experiments under the open-set personalization setting underscore the efficacy of our method.

\end{abstract}

%% file: sec/1-Introduction.tex
\section{Introduction}

Imagine a future intelligent assistant that, through prolonged interactions, can not only remember you, but also gradually uncover and internalize your personal entities that are never specified in advance: your blue cup, and your yellow puppy Timi.
When the assistant ``sees'' these entities in later conversations, it can visually recognize them and generate ``personalized responses'' that are grounded in the appropriate contexts, rather than \textit{one-size-fits-all} answers.

All these imaginations are closely tied to the research field of multimodal large language model (MLLM) personalization~\cite{alaluf2024myvlm,nguyen2024yollava,kim2025mmpb,hong2026tameing}, which aims to enable general-purpose MLLMs (e.g., GPT-5~\cite{gpt52025openai}, Qwen3.5~\cite{qwen3.5}) to understand user-specific concepts, also referred to as ``personalized concepts''\footnote{In this work, we use the terms ``concept'', ``personalized concept'', and ``user-specific concept'' interchangeably.} (e.g., Bob's dog), instead of treating them as generic objects.
Most existing studies, such as MyVLM~\cite{alaluf2024myvlm} and R2P~\cite{das2025training}, focus on the capability of MLLMs in visual identification of personalized concepts, followed by a simple substitution of the corresponding concept in the output generation.
Nevertheless, such static applications are \textit{overly simplified} and fail to cater to real-world demands, where user-MLLM interactions are prolonged and dynamic, and dialogues become more complex due to extended contexts.
As a remedy, TAME~\cite{hong2026tameing} explored long-context personalized dialogues with MLLMs by modeling long- and short-term attribute shifts of each concept over time, enabling the personalization experience to progressively improve over extended human-MLLM interaction.

\begin{figure}[t]
\centering
\includegraphics[width=0.97\linewidth]{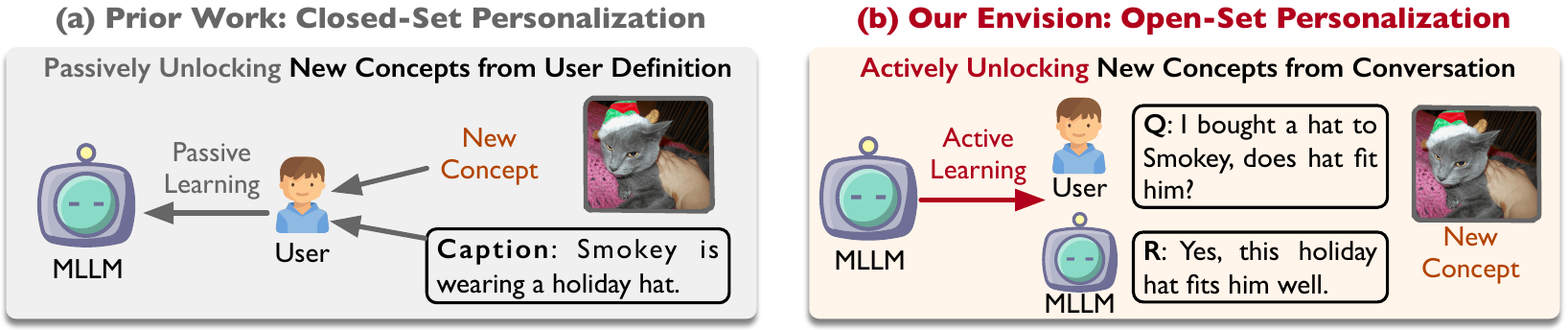}
\caption{Paradigm comparison between (a) Closed-Set Personalization, where the MLLM passively acquires novel concepts from user provision, and (b) Open-Set Personalization, where the MLLM actively grasps new concepts from everyday interactions.
}
\label{fig:intro}
\vspace{-5mm}
\end{figure}

However, all existing methods, including TAME, can only operate under the \textit{closed-set personalization setting}:
the MLLM is personalized on a fixed and static set of user-specific concepts that are manually predefined by the user.
In real-world scenarios, users frequently encounter new personalized concepts~\cite{hao2024rap,das2025training}, and this rigidity severely undermines the flexibility of existing methods, which require users to frequently register the ``novel personalized concepts'' for the MLLM by hand (cf.~\Cref{fig:intro}(a)).
In contrast, an ideal MLLM assistant should \textit{gradually unlock more and more novel personalized concepts automatically} during continued human–MLLM interaction, while progressively refining the modeling of previously known concepts. 
We term this capability \textbf{Open-Set MLLM Personalization}, an appealing property for MLLM assistants in the dynamic world (cf.~\Cref{fig:intro}(b)).

To achieve open-set personalization, the core challenge lies in enabling the MLLM assistant to reliably \textit{resolve} newly emerging personalized concepts with existing ones over extended, long-context interactions.
On one hand, when novel personalized concepts are misidentified as existing ones, the assistant may \textit{accumulate erroneous contexts} for the existing concepts.
On the other hand, misclassifying an existing user-specific concept as novel not only prevents the assistant from leveraging accumulated contexts during generation, but also leads to \textit{concept fission}, where an existing concept is erroneously forked into two or more concepts.

To address these two symmetric problems, we draw 
upon the insight that 
user-specific concepts (e.g., visual appearance)
evolve over time.
As a result, the past dialogues that are \textit{temporally proximal to the current interaction} 
often provide more reliable cues for identification than distant history.
Accordingly, when comparing a newly appearing concept with those grasped from past dialogues, more recent memories should be emphasized for robust concept maintenance.

To this end, we propose \textbf{\M}, the first \textbf{O}pen-set, long-context personalized MLLM assistant that can automatically \textbf{TAME} novel user-specific concepts as they appear during ongoing interactions,
beyond a predefined closed-set of concepts. 
\M endows the MLLM assistant with a novel training-free Concept Reconciliation Skill (\textbf{CR Skill})
that mediates between incoming concepts and stored knowledge.
Specifically, the CR Skill equips the MLLM with a new Time-Aware Dynamic Memory (\textbf{TAD Memory}), which stores personalized concepts along with their associated attributes extracted from past interactions, where each attribute is associated with a \textit{temporal marker} to reflect when it was observed.
Subsequently, when a concept is involved in the current interaction, the CR Skill retrieves candidate relevant concepts from the TAD Memory by considering concept attributes in a \textit{temporally weighted manner}, prioritizing memories closer to the current interaction.
These retrieved candidates, together with the current user query and recent dialogue contexts, are jointly reasoned by the MLLM to determine whether the concept aligns with an existing one or represents a new concept, enabling the assistant to effectively grasp novel user-specific concepts while continuously refining unlocked ones.
The CR Skill is designed to be flexible and can be readily integrated into existing personalized MLLM assistants (e.g., TAME) in a \textit{plug-and-play} way, granting them open-set personalization capability in the dynamic world.
We list our contributions as follows:
\begin{itemize} [leftmargin=*, topsep=2pt, partopsep=0pt, itemsep=0pt]

\item 
We formalize the problem of Open-Set MLLM Personalization, moving beyond static concept dictionaries to enable continuous, autonomous concept acquisition in dynamic environments.

\item 
We propose \M, the first open-set, long-context personalized MLLM assistant.
\M introduces a novel plug-and-play CR Skill that can be integrated into existing MLLM assistants to activate their open-set capability, automatically managing the lifecycle of personalized concepts by detecting the new and refining the old.

\item 
Experiments on both existing LCMP-H and LCMP-O, a dedicated benchmark for open-set MLLM personalization newly constructed in this study, showcase the efficacy of \M.
Code is provided at \url{https://github.com/ronpay/TAME-O}.

\end{itemize}

%% file: sec/2-Related-Work.tex
\section{Related Work}
\label{sec:related-work}
\textbf{MLLM Personalization.}
MLLMs have been widely adopted as assistants, while most mainstream models still operate under a one-size-fits-all paradigm, treating all users as interchangeable and failing to adapt to individual identities, preferences, or interaction histories~\cite{kim2025mmpb,gpt52025openai,yang2025qwen3,an2025unictokens,nguyen2025yo,nie2026personavlm}.
MyVLM~\cite{alaluf2024myvlm} and Yo'LLaVA~\cite{nguyen2024yollava} pioneered the concept of MLLM personalization, which enables an MLLM with generic knowledge to respond to user visual queries in a personalized manner.
They achieved personalization by training a set of parameters to visually memorize each concept~\cite{an2024mcllava,an2026concept}.
To alleviate the frequent retraining for each newly emerging concept, studies such as RAP~\cite{hao2024rap}, PVIT~\cite{pi2025personalized}, R2P~\cite{das2025training}, and PLVM~\cite{pham2025plvm} explored memory-based or tuning-free mechanisms, where a set of user-predefined personalized concepts is first stored in memory, and an incoming concept is then treated as a query to visually match against these stored concepts for personalized response generation.
However, all these methods primarily focus on the visual identification of personalized concepts, that is, recognizing an object in an image as user-specific and replacing a generic object reference with the corresponding personalized concept name for captioning or single-turn question-answering.
Such overly simplified settings fail to meet real-world requirements, where human–MLLM interactions are persistent and highly contextualized, and personalization must account for complex dependencies from extended interaction history rather than isolated, static inputs.
To remedy such limitation, TAME~\cite{hong2026tameing} proposed the first long-context MLLM personalized assistant, which fosters MLLM to capture the variations in both long- and short-term attributes of user-specific concepts.
Nevertheless, existing methods (including TAME) are limited to only responding to a \textit{manually} predefined set of user-specific concepts and fail to adapt to dynamic real-world settings.
In contrast, we propose \mbox{\M}, which activates Open-Set Personalization capability through a training-free, plug-and-play Concept Reconciliation Skill (CR Skill), which effectively and autonomously unlocks novel personalized concepts while avoiding potential conflicts with previously ``memorized'' ones.
The additional literature reviews for memory-augmented LLM and open-set recognition are in~\Cref{app:additional-literature-review}.

%% file: sec/3-Methodology.tex
\section{Methodology: \M}
\label{sec:method}

\noindent \textbf{Revisiting TAME.}
We begin by re-formalizing the pipeline of TAME~\cite{hong2026tameing}, the first long-context personalized MLLM assistant.
TAME considers a continuous stream of human–MLLM interactions:
$\mathcal{S} = \langle \mathcal{Q}_1, \mathcal{Q}_2, \ldots,\mathcal{Q}_i, \ldots\rangle$,
where each $\mathcal{Q}_i$ denotes an atomic single-turn vision–language dialogue occurring at time step $i$, consisting of a user query $q_i$ and the corresponding assistant response $a_i$.
When responding to $q_i$, the MLLM assistant is required to leverage the potential contexts from the past dialogues tied to the personalized concept appearing in $\mathcal{Q}_i$: $a_i = \Phi_\text{MLLM}(q_i,\pi(\langle \mathcal{Q}_1, \mathcal{Q}_2,\ldots,  \mathcal{Q}_{i-1}\rangle))$, where $\pi(\cdot)$ denotes a context extraction strategy that filters out relevant information for current concept from the interaction history.
To achieve this, TAME endows the assistant with double memories, a static memory $\mathcal{M}_S$ and a dynamic memory $\mathcal{M}_D$, which are designed to maintain the long-term and short-term attributes of personalized concepts from past dialogues, respectively.
Each memory stores items as a tuple $(\texttt{Cid},\texttt{Attr})$, where $\texttt{Cid}$ is the concept identifier that is used to uniquely identify a personalized concept, while $\texttt{Attr}$ denotes an attribute extracted from a specific dialogue turn.
Subsequently, when facing an incoming personalized concept, TAME first identifies its corresponding concept identifier $\texttt{Cid}$, and then aggregates the memory items (attributes) that are marked with the same \texttt{Cid} from $\mathcal{M}_S$ and $\mathcal{M}_D$ to support personalized response generation.

However, TAME implicitly assumes a \textbf{closed-set} personalization setting, where a set of $N$ user-specific concepts is predefined and fixed throughout the interaction: $\mathcal{C} = \{c_1,c_2,\cdots,c_N\}$, where $\mathcal{C}$ is the static concept set.
When encountering a novel concept $c_{N+1}$, TAME requires the user to manually initialize this concept and provide the corresponding concept identifier for proper memory management.
Consequently, such a setting is ill-suited to dynamic real-world scenarios, where new personalized concepts frequently arise, and requiring users to explicitly specify concept identifiers introduces significant overhead and undermines usability.

\noindent \textbf{Open-Set MLLM Personalization Formalization.}
We consider a more real-world setting, in which personalized concepts are not predefined by users but are instead \textbf{automatically grasped} by the MLLM assistant through extended interactions.
Specifically, the concept set is defined as: $\mathcal{C}_{i-1} = \{c_1,c_2,\cdots,c_{K_{i-1}}\}$, where $\mathcal{C}_{i-1}$ is the dynamic concept set at time step $i-1$ that contains $K_{i-1}$ concepts unlocked by the assistant up to the current step (empty at start).
At time step $i$, the concept referred to in the dialogue $\mathcal{Q}_{i}$ is judged to be either \textbf{novel} or \textbf{existing}.
If it is identified as a novel concept, the personalized concept set is updated by the assistant as: $\mathcal{C}_{i} = \{c_1,c_2,\cdots,c_{K_{i-1}},c_{{K_i}}\}$, where $c_{{K_i}}$ is the new concept appearing in $\mathcal{Q}_{i}$.

\begin{figure*}[t]
\centering
\includegraphics[width=\linewidth]{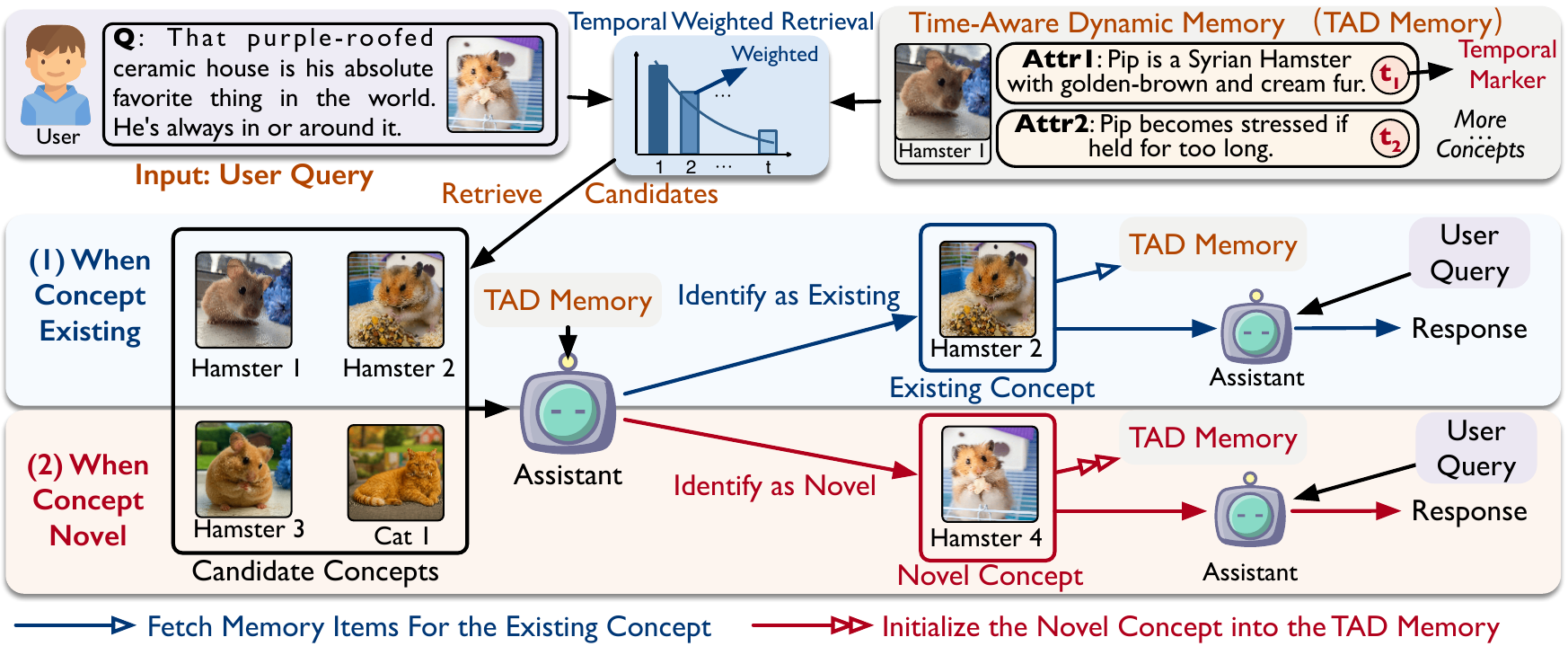}
\vspace{-4mm}
\caption{Paradigm of \M. When receiving a new dialogue, the CR skill first retrieves candidate personalized concepts from the TAD Memory in a temporally weighted manner.
The retrieved candidates are then provided to the MLLM, together with the user query and recent dialogue context, to determine whether the referenced concept is existing or novel, achieving open-set personalization.
}
\label{fig:method}
\vspace{-2mm}
\end{figure*}

\noindent \textbf{Paradigm of \M.}
To address this challenging setting, \M provides existing personalized assistants with a novel training-free and plug-and-play Concept Reconciliation Skill (CR Skill), which handles novel concepts while mitigating potential conflicts with existing ones.
Specifically, \M first equips the MLLM assistant with a Time-Aware Dynamic Memory (TAD Memory), which captures variations of personalized concepts in a temporally annotated manner (cf.~\Cref{subsec:TAD-Memory}).
Subsequently, when a new interaction occurs, the CR Skill evaluates the incoming concept against memory items using a temporally weighted matching strategy, and provides a set of candidate concepts to the MLLM for final resolution of this concept (cf.~\Cref{subsec:CR-Skill}).
We further provide an example of integrating the CR Skill into the existing personalized assistant TAME (cf.~\Cref{subsec:plug}).
Finally, we discuss potential concept fission issues and present corresponding mitigation solutions (cf.~\Cref{subsec:concept-fission}).
The framework of \M is in~\Cref{fig:method}.

\subsection{Time-Aware Dynamic Memory}
\label{subsec:TAD-Memory}

We consider that after time step $i-1$, the personalized concepts that have been identified by the MLLM assistant are represented as a dynamic concept set
$\mathcal{C}_{i-1} = \{c_1, c_2, \ldots, c_{K_{i-1}}\},$
where $K_{i-1}$ denotes the number of personalized concepts discovered up to time step $i-1$.
Given a new dialogue $\mathcal{Q}_{i}$, a simple approach to determining whether a novel personalized concept appears is to compare the information of the concept extracted from $a_i$ against the grasped concepts in memory.
However, in the long-context interactions, the information (also referred to as attributes) of each concept often varies over time~\cite{hong2026tameing}.
Consequently, early attributes of a concept may become outdated and unfaithful for depicting such concept, whereas more recent attributes can better reflect its current state.

Inspired by this observation, we introduce a novel training-free Concept Reconciliation Skill (\textbf{CR Skill}), which first endows the MLLM assistant with a new Time-Aware Dynamic Memory (\textbf{TAD Memory}) to store each attribute in a temporally-annotated manner.
Specifically, taking completing a dialogue $\mathcal{Q}_k$ and storing the newly acquired information tied to the corresponding personalized concept as an example, the TAD Memory $\mathcal{M}_T$ operates as follows:
\begin{align}
(\texttt{Cid}_k, \texttt{Attr}_k&, \delta_k)
=
\Phi\!\left(
\texttt{Cid}_k, \mathcal{Q}_k, \mathcal{M}_T, \mathcal{P}_{u}
\right), \\
\mathcal{M}_T 
\leftarrow 
\Psi\!&\left( 
\mathcal{M}_T,\texttt{Cid}_k, \texttt{Attr}_k, \delta_k, t_k 
\right),
\end{align}
where $\texttt{Cid}_k$ is the identifier of the concept referred to in $\mathcal{Q}_k$ (its acquisition is discussed in the \textbf{next section}), 
$\texttt{Attr}_k$ is the attribute information extracted from $\mathcal{Q}_k$,
$\delta_k$ is the memory execution indicator determined by the MLLM $\Phi(\cdot)$ for this attribute $\texttt{Attr}_k$ that follows the update actions adopted in TAME (i.e., add or modify an attribute),
$\mathcal{P}_{u}$ is the prompt for memory update,
and $\Psi(\cdot)$ is a memory update operation that updates the TAD Memory according to the indicator $\delta_k$.
$t_k$ is the temporal marker for $\texttt{Attr}_k$, which is separately maintained for each concept and set to 0 for the first stored memory item (attribute).
It is incremented by 1 and assigned to each new attribute that is added or updated.
Consequently, TAD Memory stores each attribute in a triple tuple: $(\texttt{Cid}_k,\texttt{Attr}_k,t_k)$.
If $\texttt{Cid}_k$ is determined to be novel, then the concept set is updated to $\mathcal{C}_{i} = \{c_1,c_2,\cdots,c_{K_{i-1}},c_{{K_i}}\}$.

\subsection{Open-Set Concept Reconciliation}
\label{subsec:CR-Skill}
When starting a new dialogue $\mathcal{Q}_{i}$, the assistant is required to identify the concept appearing in user query $q_{i}$, resolving whether it is novel or existing and giving its appropriate identifier.
Specifically, the CR Skill first retrieves a set of candidate personalized concepts by comparing the visual~\footnote{We first perform a visual grounding to identify the concept in the query (cf. Implementation Details)} and textual information extracted from $q_{i}$ with the information of each concept $c_k$ stored in the TAD Memory in a \textbf{temporally weighted manner}, emphasizing the memory items that are closer to the $q_{i}$:
\begin{align}
S(c_k \mid q_{i})
=&
\frac{1}{Z_k}
\sum_{j=1}^{N_k}
\exp\!\left(
-\lambda \cdot \frac{t_k^{\max} - t_{k}^{(j)}}{N_k}
\right)
\cdot s_{k}^{(j)},
\\
s_{k}^{(j)}
=\cos (\phi(&\mathcal{T}_{i}),\phi(\mathcal{T}_{k}^{(j)})) + \cos (\phi(\mathcal{V}_{i}),\phi(\mathcal{V}_{k}^{(j)})).
\end{align}
Here, $S(c_k \mid q_{i})$ denotes the temporally weighted similarity score between the information extracted from $q_{i}$ and the $N_k$ historical memory items associated with the personalized concept $c_k$ in $\mathcal{M}_T$: $\{(\mathcal{T}_{k}^{(j)}, \mathcal{V}_{k}^{(j)}, t_{k}^{(j)})\}_{j=1}^{N_k}$.
$\mathcal{T}_{i}$ and $\mathcal{V}_{i}$ represent the textual and visual information extracted from the incoming user query $q_{i}$, while $(\mathcal{T}_{k}^{(j)}, \mathcal{V}_{k}^{(j)})$ correspond to the textual and visual attributes stored in the $j$-th memory item of concept $c_k$.
$t_{k}^{(j)}$ is the temporal marker for that item, and $t_k^{\max} = \max_j t_{k}^{(j)}$ denotes the most recent temporal marker of attributes for concept $c_k$.
$s_{k}^{(j)}$ is the similarity score between the current query and the $j$-th memory item, computed using cosine similarity in the embedding space defined by $\phi(\cdot)$.
$\lambda$ is the temporal decay coefficient, and $Z_k$ is the normalization term ensuring a scaled concept-level similarity score, defined as:
\begin{equation}
    Z_k = \sum_{j=1}^{N_k} \exp\!\left(-\lambda \cdot \frac{t_k^{\max} - t_{k}^{(j)}}{N_k}\right).
\end{equation}

Subsequently, the CR skill selects the Top-$K$ candidate personalized concepts $\mathcal{C}_K$ with the highest similarity scores from the grasped concept set $\mathcal{C}_{i-1}$ after time step $i-1$.
For all candidates, the assistant feeds their most recent memory items, together with the current query 
$q_i$ and the most recent 
$W$ dialogue turns, into the MLLM to determine whether the referred concept corresponds to an existing personalized concept or a novel one:
\begin{equation}
\label{eq:5}
\texttt{Cid}^{*}
=
\Phi\left(
\{\mathcal{M}_{k}^{\text{r}}\}_{c_k \in \mathcal{C}_K}, q_i,
\langle \mathcal{Q}_{i-W}, \ldots, \mathcal{Q}_{i-1} \rangle, \mathcal{P}_{r}
\right).
\end{equation}
Here, $\texttt{Cid}^{*}$ denotes the concept identifier predicted by the MLLM for the personalized concept referred to in the incoming dialogue $\mathcal{Q}_{i}$. 
The output $\texttt{Cid}^{*}$ can either correspond to an existing concept identifier in the candidate concept set $\mathcal{C}_K$ or indicate a novel concept, which will be initialized with a new identifier.
$\mathcal{M}_{k}^{\text{r}}$ represents the recent $R$ memory items associated with concept $c_k$, which reflects its latest attribute state.
$q_i$ denotes the current user query in the dialogue turn, and $\langle \mathcal{Q}_{i-W}, \ldots, \mathcal{Q}_{i-1} \rangle$ denotes the most recent $W$ dialogue turns serving as recent conversational context.
$\mathcal{P}_{r}$ is the prompt that instructs the MLLM to perform concept reconciliation and to output either an \textbf{existing concept identifier} or a \textbf{novel one}.
Consequently, the CR skill completes the concept reconciliation.

\subsection{Plug-and-Play Operation}
\label{subsec:plug}
We take the integration of \M (i.e., the training-free CR Skill) into TAME as an example. 
Specifically, we retain TAME's static memory $\mathcal{M}_{S}$ while replacing its dynamic memory $\mathcal{M}_{D}$ with TAD Memory $\mathcal{M}_{T}$.
After obtaining the determined concept identifier $\texttt{Cid}^{*}$ in $\mathcal{Q}_{i}$ in Eq.~(\ref{eq:5}), if the concept is \textbf{an existing one}, the assistant leverages this identifier to locate relevant memory items from both memories, and aggregates the corresponding information for personalized response generation (following the original TAME pipeline).
If the concept is determined to be \textbf{novel}, a new identifier will be assigned to this concept and stored in the TAD Memory, therefore \textbf{unlocking new concept automatically}.
Notably, when integrated with TAME, we also incorporate information from the static memory for candidate concept retrieval.
The final similarity score for each candidate concept is obtained by averaging the similarity scores from the static memory and the TAD Memory.

\begin{figure}[t]
\centering
\includegraphics[width=0.75\linewidth]{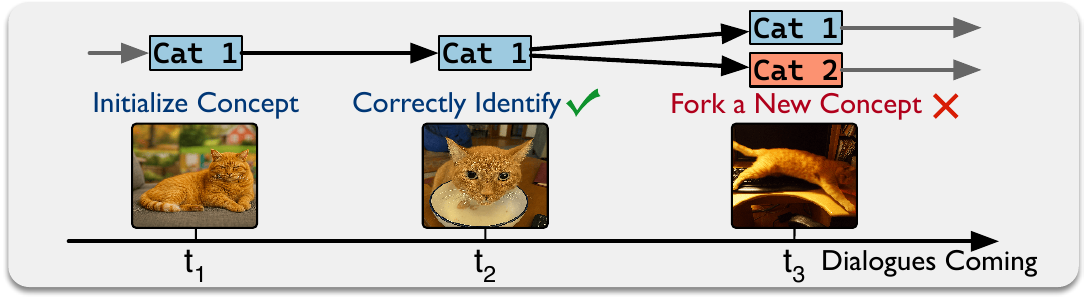}
\vspace{-1mm}
\caption{A concept fission example for concept \texttt{Cat 1}.
}
\vspace{-3mm}
\label{fig:fission}

\end{figure}

\subsection{Discussion of Concept Fission and Its Mitigation}
\label{subsec:concept-fission}
Although the CR Skill effectively resolves novel and existing concepts during ongoing interactions, it may still misidentify an existing concept as a novel one.
In such cases, the original concept unit is unintentionally forked into multiple concept units with different identifiers, a phenomenon we refer to as \textbf{concept fission}. 
This leads to memory chaos, where no single concept unit retains complete contextual information, ultimately resulting in degraded personalization performance.
To stably mitigate this issue, we introduce a Periodic Concept Fusion Strategy (\textbf{PCF Strategy}), which is performed every fixed interval of $T$ interaction steps.
Specifically, after each interval $T$, we compute pairwise similarities between all personalized concepts based on their stored memory items, and select the Top-$L$ most similar concept pairs as candidates for fusion:
\begin{align}
S(c_m, c_n)
&=
\frac{1}{|\mathcal{M}_m|\,|\mathcal{M}_n|}
\sum_{i=1}^{|\mathcal{M}_m|}
\sum_{j=1}^{|\mathcal{M}_n|}
s_{m,n}^{(i,j)}, \label{eq:concept-pair-score} \\
s_{m,n}^{(i,j)}
= 
\cos (\phi(&\mathcal{T}_m^{(i)}),\phi(\mathcal{T}_n^{(j)})) + \cos (\phi(\mathcal{V}_m^{(i)}),\phi(\mathcal{V}_n^{(j)})).
\end{align}
Here, $S(c_m, c_n)$ is the pairwise similarity between two concepts $c_m$ and $c_n$.
$\mathcal{M}_m$ and $\mathcal{M}_n$ represent the memory sets associated with concepts $c_m$ and $c_n$, respectively.
$s_{m,n}^{(i,j)}$ is the similarity between the $i$-th memory item of concept $c_m$ and the $j$-th memory item of concept $c_n$, computed using cosine similarity in the embedding space defined by $\phi(\cdot)$.
The overall concept-level similarity $S(c_m, c_n)$ is obtained by averaging the similarities over all memory item pairs, yielding a robust measure of semantic closeness between the two concepts. 
Subsequently, the Top-$L$ concept pairs are derived:
\begin{equation}
    \mathcal{F}
=
\operatorname{Top}\text{-}L
\Big\{
(c_m, c_n)\ \big|\ S(c_m, c_n)
\Big\}_{m \neq n}.
\end{equation}
Afterward, we further introduce a similarity threshold $\tau$ to identify the most potentially forked concept pairs in $\mathcal{F}$,
and leverage the MLLM assistant with prompt $\mathcal{P}_f$ to evaluate each concept pair to perform the fusion and then update the memories.
The prompts $\mathcal{P}_r$ and $\mathcal{P}_f$ are presented in~\Cref{app:prompt}, while the pseudocode is provided in~\Cref{app:pseudocode}.
The complexity analysis of \M is in~\Cref{app:complexity}.

%% file: sec/4-Experiments.tex
\section{Experiments}
\label{sec:experiments}

We provide a brief version of the experimental settings, with a detailed version in~\Cref{app:experimental_setup}.

\noindent \textbf{Benchmarks.}
Since open-set personalization is inherently intertwined with long-context dialogues, we first adopt an existing long-context personalized MLLM benchmark \textbf{LCMP-H}~\cite{hong2026tameing}, which consists of 30 personalized concepts (Pet (10), Object (10), and Person (10)), 420 historical dialogue turns, 720 personalized images, and 180 hard questions.
In addition, we introduce a new benchmark tailored for long-context open-set personalization, termed \textbf{LCMP-O}. 
Compared with LCMP-H, LCMP-O poses a more challenging open-set scenario by incorporating a \textit{larger number} of personalized concepts (45 concepts) with more \textit{visually similar appearances}. 
Specifically, its concepts are evenly split into Pet (15), Object (15), and Person (15), where pets and objects are organized into groups of three concepts sharing the same breed or object class, and persons are selected with similar ages and an approximately 1:1 gender ratio.
Following LCMP~\cite{hong2026tameing}, we generate 10 historical dialogue turns for each concept to reveal its long-term attributes and short-term events, as well as 5 hard questions requiring reasoning over both types of attributes.
Overall, LCMP-O contains 450 historical dialogue turns, 675 personalized images, and 225 hard questions, with more details in~\Cref{app:lcmp-o}.

\noindent \textbf{Open-Set Personalization Setup.}
To reflect the open-set setting, we adopt a \textbf{concept-incremental evaluation protocol}. 
Specifically, personalized concepts are introduced to MLLMs sequentially rather than all at once. 
For each concept, the MLLM is first exposed to its associated historical dialogues in a \textit{temporally sequential manner}, one dialogue turn at a time step~\footnote{We define time steps globally across concepts, where each dialogue turn constitutes a time step.}. 
After all dialogue turns for the concept have been processed, the MLLM is instantly evaluated on the corresponding QA set tied to that concept.
Moreover, in contrast to~\cite{hong2026tameing}, the concept identifiers of historical dialogues under the open-set setting \textbf{are not} provided to the assistant.
Instead, the assistant is required to autonomously and incrementally unlock new concepts.

\noindent \textbf{Baselines.}
We first compare \M against several state-unaware MLLM personalization baselines, including Backbone (the native MLLM without personalization adaptation), MyVLM~\cite{alaluf2024myvlm}, Yo’LLaVA~\cite{nguyen2024yollava}, RAP~\cite{hao2024rap}, PeKit~\cite{seifi2025personalization}, and R2P~\cite{das2025training}.
Notably, we do not apply the open-set setting to these baselines, as their state-unaware design prevents them from directly supporting unlocking novel concepts automatically across interactions.
In addition, we construct a new open-set personalization baseline, TAME$^{\dagger}$, by extending the long-context baseline TAME~\cite{hong2026tameing} with a simple retrieval-based concept unlocking and resolution mechanism. 
Specifically, TAME$^{\dagger}$ retrieves candidate concepts from the memory and treats the Top-1 retrieved concept as the matched existing concept, if its similarity exceeds a predefined threshold.
Otherwise, the concept is considered novel.
We further experiment with TAME$^{\dagger}$ w/ MLLM decision (i.e., the w/o TAD Memory variant in~\Cref{tab:ablation}).

\noindent \textbf{Evaluation Metrics.}
Following LCMP, we employ evaluation metrics: \textbf{(1)} Accuracy of Choice questions (\textbf{ACC-C}), \textbf{(2)} Accuracy of Free-text questions (\textbf{ACC-F}), \textbf{(3)} Scoring Point Rate Long-term attributes (\textbf{SPR-L}), and \textbf{(4)} Scoring Point Rate Short-term attributes (\textbf{SPR-S}), which measure the proportion of responses that explicitly reference the relevant attribute variation of the personalized concept.
Moreover, we propose \textbf{(5)} Concept Reconciliation Accuracy (\textbf{CRA}) as a new metric for evaluating the open-set personalization capability.
Formally, CRA can be defined as:
\begin{equation}
\mathrm{CRA}
=
\frac{1}{|\mathcal{H}|}
\sum_{i=1}^{|\mathcal{H}|}
\mathbb{I}\!\left(
\hat{\texttt{Cid}}_i = \texttt{Cid}_i
\right),
\end{equation}
where $\mathcal{H}$ is the set of historical dialogues, $\mathbb{I}(\cdot)$ is the indicator function that returns 1 if the prediction is correct and 0 otherwise, $\hat{\texttt{Cid}}_i$ is the concept identifier determined by the MLLM assistant in $i$-th historical dialogue.
$\texttt{Cid}_i$ denotes the ground-truth concept identifier for the $i$-th dialogue.
If the referred concept is an existing one (already grasped), $\texttt{Cid}_i$ corresponds to the identifier of that concept. 
If it is novel, $\hat{\texttt{Cid}}_i$ is expected to be a previously unused identifier.
If an existing concept has been forked into multiple identifiers (concept fission), any predicted identifier $\hat{\texttt{Cid}}_i$ that corresponds to one of the forked identifiers is treated as a correct match.
Moreover, we propose the metric \textbf{(6)} Concept Fission Rate (\textbf{CFR}) to measure the fission rate of the concepts at different time steps $t$:
\begin{equation}
\label{eq:10}
\mathrm{CFR}(t)
=
\frac{
\sum_{c_i \in \mathcal{C}_t}
\max\!\left(0,\ |\hat{\mathcal{I}}_{c_i}^{(t)}| - 1 \right)
}{
\max (t -
|\mathcal{C}_t|, \epsilon)
},
\end{equation}
where $\mathcal{C}_t = \{c_1, \ldots, c_{N_t}\}$ denotes the set of ground-truth personalized concepts that exist at time step $t$, $\hat{\mathcal{I}}_{c_i}^{(t)}$ is the set of concept identifiers derived by the MLLM assistant at time step $t$ associated with the ground-truth concept $c_i$, which contains the correct identifier and may unintentionally include forked identifiers introduced due to concept fission.
As a result, $\mathrm{CFR}(t)$ is designed to characterize the temporal evolution of concept fission degree during long-term interactions.

\noindent \textbf{Implementation Details.}
We adopt the MLLM \texttt{Qwen3.5-9B}~\cite{qwen3.5} as our main backbone, while also evaluating \M on another MLLM \texttt{InternVL3.5-8B}~\cite{wang2025internvl} for generalizability.
Since \M is plug-and-play, we adopt TAME as the base personalized assistant for evaluation.
In addition, we also evaluate its scalability on other MLLM personalization methods to unlock their open-set capability (cf. Appendix~\ref{app:scalability}).
When visually grounding the concept in user's query, we first prompt the MLLM to extract open-set category descriptions from the query, and then use the extracted categories to perform grounding on the query image with \texttt{Grounding-DINO}~\cite{liu2024grounding}. 
If no concept is grounded, we provide the whole image to the MLLMs for concept reconciliation.
In addition, we adopt \texttt{Qwen3-VL-Embedding-2B}~\cite{li2026qwen3vlembedding} to generate embeddings for both visual and textual inputs.

\input{sec/table-1}

\begin{figure}[t]
    \centering
    \begin{subfigure}[b]{0.49\columnwidth}
    \centering
    \includegraphics[width=\columnwidth]{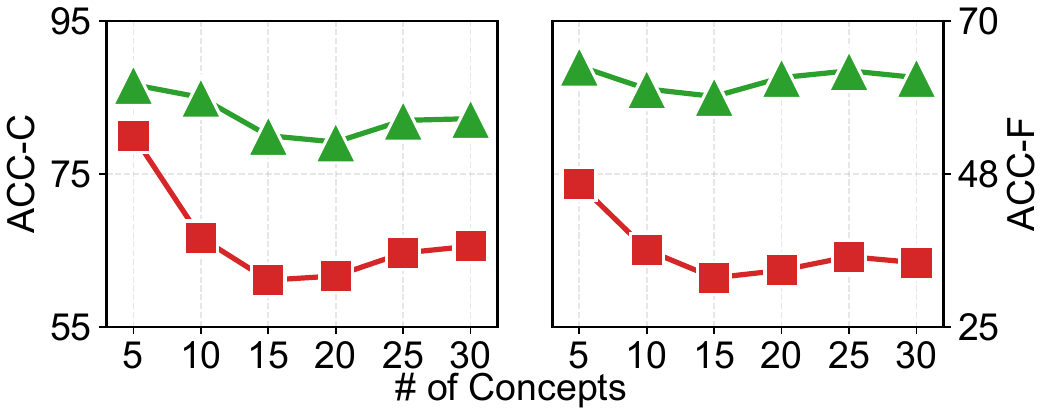}
    \vspace{-6mm}
    \caption{LCMP-H.}
    \end{subfigure}
    \hfill
    \begin{subfigure}[b]{0.49\columnwidth}
    \centering
    \includegraphics[width=\columnwidth]{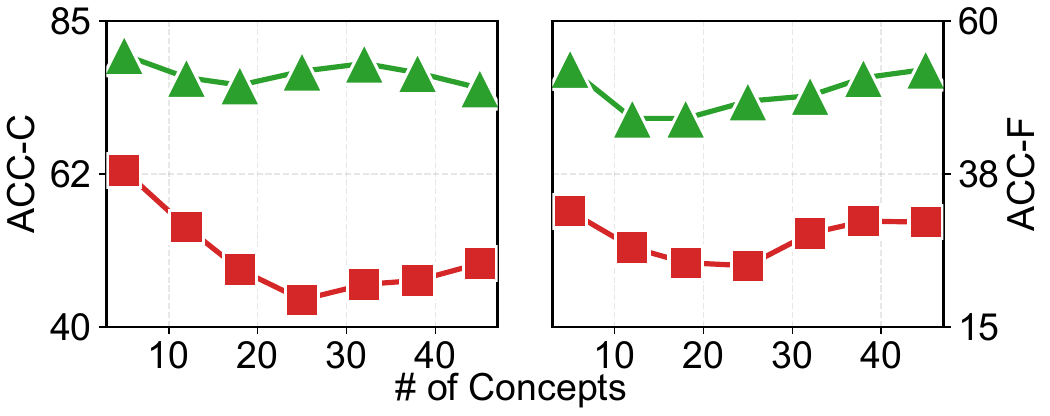}
    \vspace{-6mm}
    \caption{LCMP-O.}
    \end{subfigure}
    \vspace{-2mm}
    \caption{Performance comparison between TAME$^{\dagger}$ and \M with respect to a continually increasing number of concepts. {\color{darkred}$\blacksquare$}~denotes TAME$^{\dagger}$, while {\color{darkgreen}$\blacktriangle$}~denotes \M.}
    \label{fig:continually-improve-concept}
    \vspace{-5mm}
\end{figure}

\begin{table}[!t]
    \centering
    \caption{Ablation study on core components in \M with Qwen3.5-9B.}
    \label{tab:ablation}
        \setlength{\tabcolsep}{5pt}
         \renewcommand{\arraystretch}{1.15}
    \resizebox{\linewidth}{!}{
    \begin{tabular}{lc|ccccc|ccccc}
        \shline
         & & \multicolumn{5}{c|}{\textbf{LCMP-H}} & \multicolumn{5}{c}{\textbf{LCMP-O}}\\
       \textbf{Module} &  \textbf{Variant} & \textbf{ACC-C} & \textbf{ACC-F} &\textbf{SPR-L} & \textbf{SPR-S} & \textbf{CRA} & \textbf{ACC-C} & \textbf{ACC-F} &\textbf{SPR-L} & \textbf{SPR-S} & \textbf{CRA} \\
        \hline
     \multirow{3}{*}{\makecell[c]{CR Skill}} 
    & w/o TAD Memory  & 77.78 & 57.22 & 60.56 & 68.89 & 69.67 & 70.67 & 48.44 & 68.44 & 53.33 & 66.59 \\ 
    & w/o MLLM Decision & 72.78 & 48.89 & 40.00 & 52.22 & 51.13 & 46.22 & 37.78 & 52.44 & 29.33 & 46.33 \\
    & w/o CR Skill & 66.11 & 38.33 & 24.44 & 45.56 & 36.84 & 52.44 & 34.67 & 50.67 & 29.33 & 33.41 \\
    \hline
    PCF Strategy & w/o PCF Strategy & 78.89 & 59.44 & 61.11 & 67.22 & 69.92 & 73.78 & 47.56 & 70.22 & 54.67 & 68.15 \\ \hline
    \rowcolor{gray!10}
    \textbf{\M} & \textbf{ALL}  & \textbf{82.22} & \textbf{61.67} & \textbf{64.44} & \textbf{70.00} & \textbf{72.18} & \textbf{75.11} & \textbf{52.89} & \textbf{72.00} & \textbf{59.11} & \textbf{71.49} \\ 
    \shline
    \end{tabular}
    }
    \vspace{-3mm}
\end{table}

\subsection{Open-Set Personalization Performance}
From the results in~\Cref{tab:main-result}, we observe that:
\textbf{(O1)} 
Training-based methods, such as MyVLM and Yo’LLaVA, perform poorly on both benchmarks, as encoding long-context personalized information with a limited number of lightweight parameters is insufficient to capture extended interaction histories. 
In contrast, RAG-based, training-free methods such as R2P and PeKit achieve stronger performance by storing personalized concepts in external memory while keeping model parameters fixed. 
However, all these baselines operate under a closed-set assumption and lack the ability to automatically incorporate novel personalized concepts.
\textbf{(O2)} TAME is the SOTA baseline for long-context personalization.
However, under the open-set setting, TAME$^{\dagger}$ exhibits suboptimal performance, since simply treating all historical attributes equally fails to identify appropriate matched concepts.
Moreover, the concept fission problem further undermines the effectiveness of TAME$^{\dagger}$. 
In contrast, \M achieves the best performance for two reasons: the CR Skill accurately reconciles incoming personalized concepts by treating the stored concept attributes in a temporally weighted manner, emphasizing the most recently observed concept information.
Moreover, the PCF Strategy further stably fuses fissioned concepts, mitigating the concept fission problem and enabling the better localization of sufficient contextual information for each concept.
Furthermore, \Cref{fig:continually-improve-concept} shows that \M maintains stable and effective open-set personalization performance as the total number of concepts continually increases.

\begin{figure}[t]
    \centering
    \begin{subfigure}[b]{0.24\columnwidth}
    \centering
    \includegraphics[width=\columnwidth]{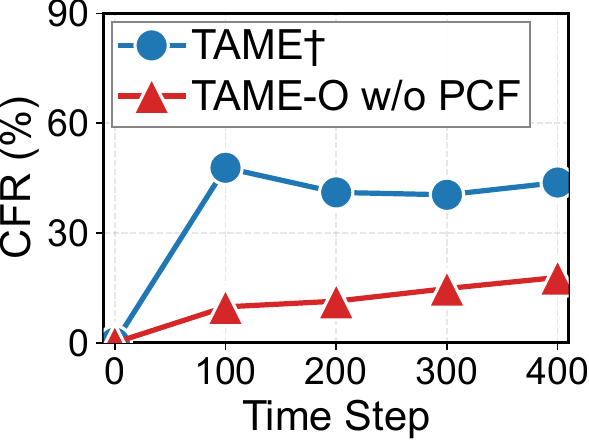}
    \vspace{-6mm}
    \caption{LCMP-H.}
    \end{subfigure}
    \hfill
    \begin{subfigure}[b]{0.24\columnwidth}
    \centering
    \includegraphics[width=\columnwidth]{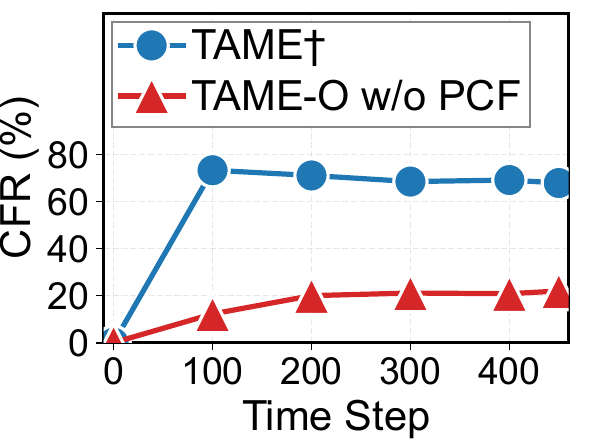}
    \vspace{-6mm}
    \caption{LCMP-O.}
    \end{subfigure}
    \hfill
    \begin{subfigure}[b]{0.24\columnwidth}
    \centering
    \includegraphics[width=\columnwidth]{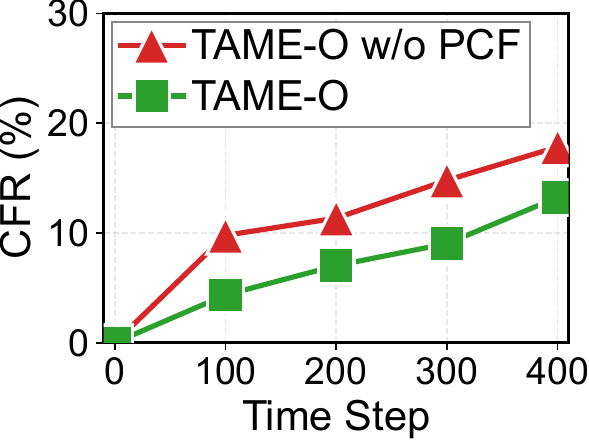}
    \vspace{-6mm}
    \caption{LCMP-H.}
    \end{subfigure}
    \hfill
    \begin{subfigure}[b]{0.24\columnwidth}
    \centering
    \includegraphics[width=\columnwidth]{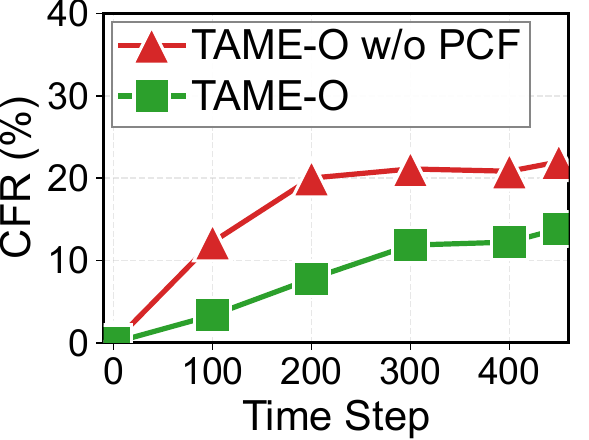}
    \vspace{-6mm}
    \caption{LCMP-O.}
    \end{subfigure}
    \caption{Comparison of CFR among TAME$^{\dagger}$, \M w/o PCF Strategy, and \M.}
    \label{fig:concept-fission}
    \vspace{-4mm}
\end{figure}

\begin{figure}[t]
    \centering
    \begin{subfigure}[b]{0.32\columnwidth}
    \centering
    \includegraphics[width=\columnwidth]{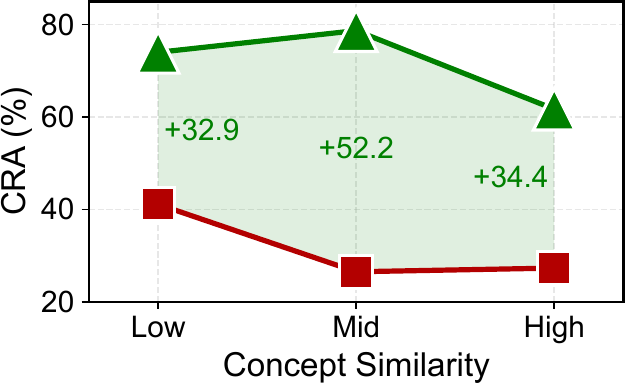}
    \vspace{-5mm}
    \caption{Reconciliation accuracy.}
    \end{subfigure}
    \hfill
    \begin{subfigure}[b]{0.32\columnwidth}
    \centering
    \includegraphics[width=\columnwidth]{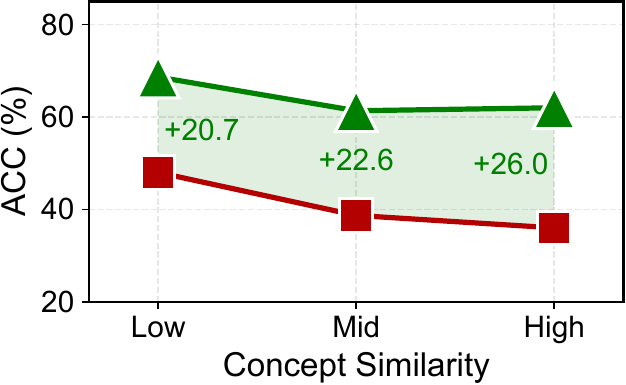}
    \vspace{-5mm}
    \caption{QA accuracy.}
    \end{subfigure}
    \hfill
    \begin{subfigure}[b]{0.32\columnwidth}
    \centering
    \includegraphics[width=\columnwidth]{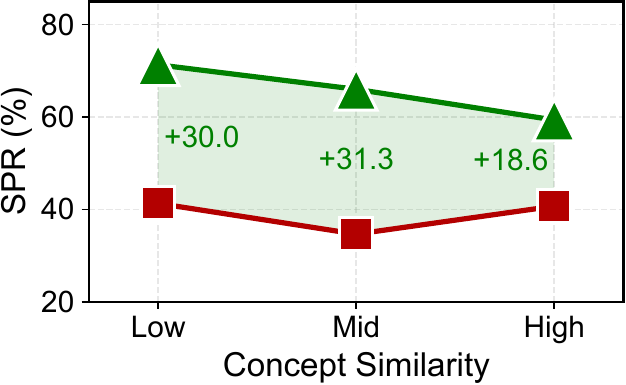}
    \vspace{-4mm}
    \caption{Scoring point rate.}
    \end{subfigure}
    \caption{Performance on LCMP-O with concepts stratified by confusability. ACC and SPR average ACC-C/ACC-F and SPR-L/SPR-S, respectively. {\color{darkred}$\blacksquare$}~denotes TAME$^{\dagger}$, while {\color{darkgreen}$\blacktriangle$}~denotes \M.}
    \label{fig:sim-bins}
    \vspace{-6mm}
\end{figure}

\subsection{Further Analysis}
\textbf{Ablative Analysis.} 
We conduct comprehensive ablations on the core components within \M, with results in~\Cref{tab:ablation}.
For \textbf{CR Skill}, we evaluate the variants: 
\textbf{(1) w/o TAD Memory}, where the Time-Aware Dynamic (TAD) Memory is reduced to a static memory and all temporal markers associated with concept attributes are removed;
\textbf{(2) w/o MLLM Decision}, where the Top-1 retrieved concept is directly treated as the matched concept if its similarity score exceeds a predefined threshold, and otherwise regarded as novel;
\textbf{(3) w/o CR Skill}, where both the TAD Memory and the MLLM-based decision mechanism are removed, and a simple time-unaware retrieval strategy is applied, with the Top-1 retrieved concept always selected as the matched existing concept if it exceeds a predefined threshold.
We observe that variants (1)--(3) exhibit clear performance drops, indicating that the accurate reconciliation of novel and existing user-specific concepts is vital for open-set personalization.
For \textbf{PCF Strategy}, we entirely remove it to design a variant: \textbf{w/o PCF Strategy}.
We observe that this variant incurs a clear performance drop, as the degree of concept fission becomes increasingly severe over time without any mitigation,
leading to memory chaos.

\textbf{Concept Fission and Fusion Analysis.}
We quantitatively evaluate the severity of the concept fission and how well \M mitigates the fission problem.
Specifically, we report the Concept Fission Rate (CFR), as defined in Eq.~(\ref{eq:10}), over time steps on the historical dialogues of LCMP-H and LCMP-O.
From~\Cref{fig:concept-fission}(a)--(b), we observe that the CFR of TAME$^{\dagger}$ rapidly reaches 47.8\% and 73.3\% on LCMP-H and LCMP-O and then remains high, while that of \M w/o PCF Strategy gradually increases.
Nevertheless, \M w/o PCF Strategy has a much lower CFR than TAME$^{\dagger}$, since the CR skill mitigates concept fission at its source by more accurately reconciling incoming open-set personalized concepts with existing ones.
From \Cref{fig:concept-fission}(c)--(d), we observe that with the PCF strategy equipped, the CFR never exceeds that without it and finally drops from 17.8\% to 13.2\% (LCMP-H) and from 22.0\% to 13.8\% (LCMP-O), indicating that the PCF strategy effectively fuses potentially forked concepts.
Moreover, we empirically observe that the PCF Strategy performs fusion with relatively high accuracy, achieving a strong fusion success rate (cf. Appendix~\ref{app:pcf-strategy} for details).

\textbf{Robustness to Concept Confusability.}
Since visually similar personalized concepts are where open-set reconciliation is most likely to fail, we further break down the performance on LCMP-O by concept confusability.
Specifically, we measure the confusability of each concept by its nearest-neighbour similarity to the other concepts, computed with the same encoder and text-plus-visual similarity score as the PCF Strategy (Eq.~(\ref{eq:concept-pair-score})), and split the 45 concepts into three equal-sized tertiles (Low, Mid, and High).
As shown in~\Cref{fig:sim-bins}, \M outperforms TAME$^{\dagger}$ on every metric in every similarity bin, with CRA gains of 32.9--52.2 points.
Moreover, \M maintains relatively stable performance as concept confusability increases, keeping its CRA above 60 even in the most confusable bin.
In contrast, the CRA of TAME$^{\dagger}$ already drops sharply at medium similarity, as its threshold-based Top-1 matching is easily misled by similar concepts.

\begin{figure}[t]
    \centering
    \begin{subfigure}[b]{0.49\columnwidth}
    \centering
    \includegraphics[width=\columnwidth]{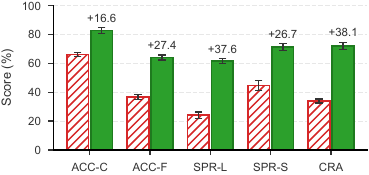}
    \vspace{-5mm}
    \caption{LCMP-H.}
    \end{subfigure}
    \hfill
    \begin{subfigure}[b]{0.49\columnwidth}
    \centering
    \includegraphics[width=\columnwidth]{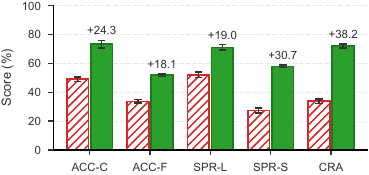}
    \vspace{-5mm}
    \caption{LCMP-O.}
    \end{subfigure}
    \vspace{-2mm}
    \caption{Performance (mean $\pm$ std) across five concept appearance orders. Numbers above the bars are the gains over TAME$^{\dagger}$. \hatchbar~denotes TAME$^{\dagger}$, while \solidbar~denotes \M.}
    \label{fig:stability-result}
    \vspace{-4mm}
\end{figure}

\textbf{Stability across Concept Appearance Orders.}
Under the open-set setup, personalized concepts emerge sequentially.
As a result, different appearance orders lead to different sets of previously acquired concepts in memory for the same incoming concept.
This, in turn, affects the retrieved candidate concepts and the difficulty of determining whether the incoming concept is existing or novel.
Here, we report the mean performance and variability of \M and TAME$^{\dagger}$ across five concept appearance orders in~\Cref{fig:stability-result}.
We observe that \M maintains stable performance across different concept appearance orders and consistently outperforms TAME$^{\dagger}$.

\begin{figure*}[t]
\centering
\includegraphics[width=\linewidth]{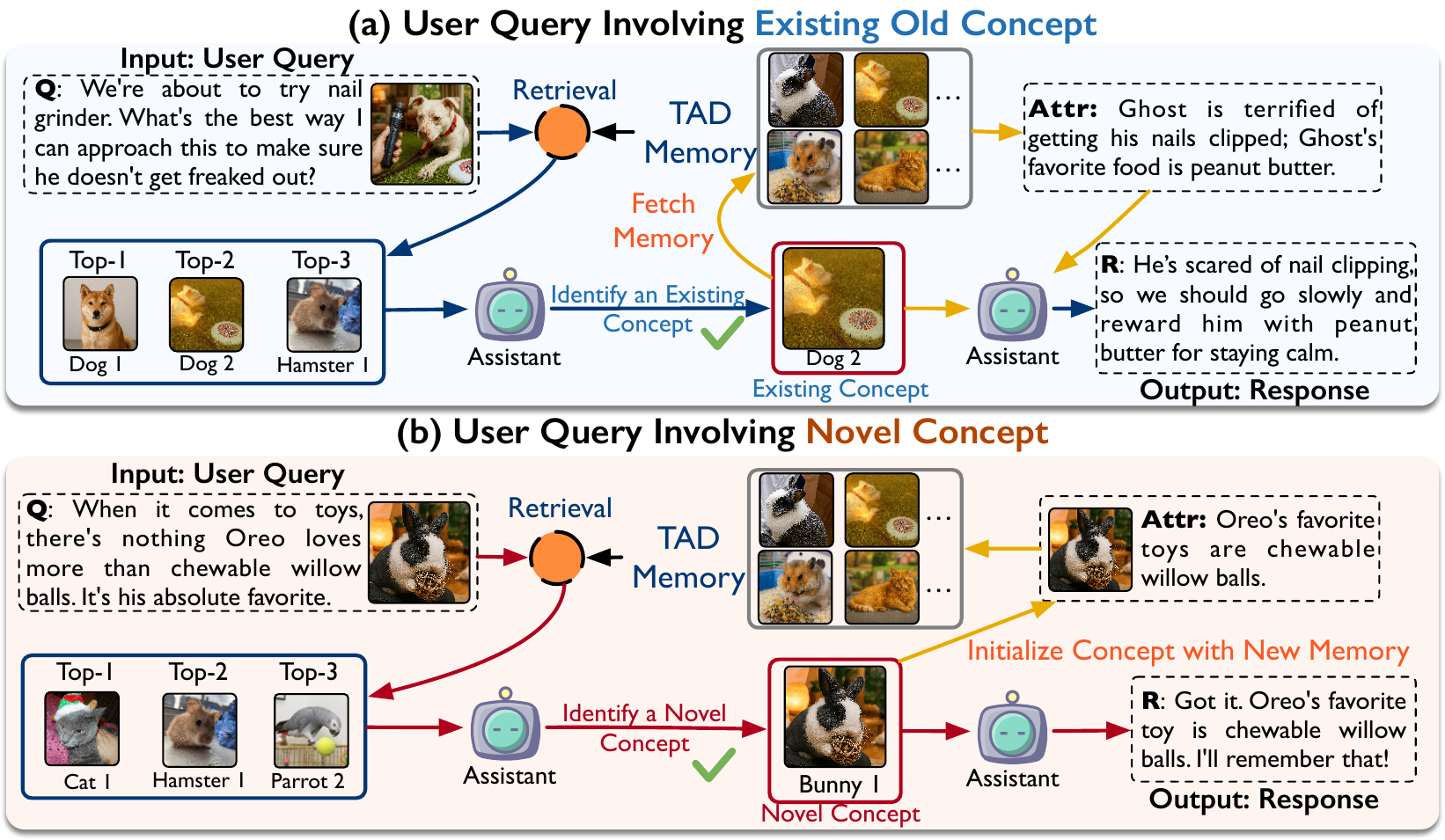}
\vspace{-3mm}
\caption{Case study on resolving an existing concept (left) and a novel concept (right).
}
\vspace{-4mm}
\label{fig:case-study}
\end{figure*}

\textbf{Case Study: Old and Novel Concept Reconciliation.}
In~\Cref{fig:case-study}, we visualize two cases to showcase the procedure of \M in tackling the open-set personalization problem.
On the top, the incoming concept is correctly identified by the CR skill as an existing one (\texttt{Dog 2}), after which the relevant memory items are retrieved for the personalized response.
On the bottom, the CR skill correctly identifies the incoming concept as novel, responds without retrieving existing memories, and unlocks this new concept with an initialized identifier \texttt{Bunny 1} in the TAD Memory.

%% file: sec/table-1.tex
\begin{table*}[t]
    \caption{Performance comparison on the LCMP-H and LCMP-O datasets. The best results are in \textbf{black bold}. 
    Higher values of all metrics indicate better performance. 
    TAME$^{\dagger}$ is the modified TAME.}
    \vspace{-1mm}
    \label{tab:main-result}
    \centering
    \renewcommand{\arraystretch}{1.02}
    \setlength{\tabcolsep}{1.5pt}
    \resizebox{\linewidth}{!}{
    \begin{tabular}{l|c|c|c|c|ccccc|ccccc}
    \shline
     \multirow{2}{*}{\textbf{Methods}} 
     & \multirow{2}{*}{\textbf{MLLM Backbone}} 
     & \multirow{2}{*}{\makecell[c]{\textbf{Training} \\ \textbf{Free}}} 
     & \multirow{2}{*}{\makecell[c]{\textbf{State} \\ \textbf{Aware}}} 
     & \multirow{2}{*}{\makecell[c]{\textbf{Open} \\ \textbf{Set}}}  
     & \multicolumn{5}{c|}{\textbf{LCMP-H}} 
     & \multicolumn{5}{c}{\textbf{LCMP-O}}\\ 
     & & & & 
     & \textbf{ACC-C} & \textbf{ACC-F} & \textbf{SPR-L} & \textbf{SPR-S} & \textbf{CRA}
     & \textbf{ACC-C} & \textbf{ACC-F} & \textbf{SPR-L} & \textbf{SPR-S} & \textbf{CRA} \\
    \hline 

   \multirow{2}{*}{\makecell[c]{Backbone}}
   & InternVL3.5-8B
    & \multirow{2}{*}{\ding{51}}
    & \multirow{2}{*}{\ding{55}}
    & \multirow{2}{*}{\ding{55}}
   & 40.00 & 20.00 & 12.78 & 26.11 & \textemdash & 32.00 & 18.67 & 16.44 & 7.11 & \textemdash \\
    & Qwen3.5-9B
   & & & & 36.11 & 20.56 & 17.22 & 32.78 & \textemdash & 38.67 & 22.22 & 29.78 & 14.67 & \textemdash \\ \hdashline

   MyVLM & LLaVA-v1.6-7B   &  \ding{55} &  \ding{55} &  \ding{55} & 40.00 & 13.33 & 8.89 & 12.22 & \textemdash & 46.67 & 21.33 & 18.22 & 6.67 & \textemdash \\ \hdashline

    Yo'LLaVA & LLaVA-v1.5-13B     &  \ding{55} &  \ding{55} &  \ding{55} & 38.89 & 9.44 & 8.89 & 17.22 & \textemdash & 56.44 & 6.67 & 20.89 & 9.78 & \textemdash \\ \hdashline

    \multirow{2}{*}{\makecell[c]{RAP}}
    & LLaVA-v1.5-13B
    & \multirow{2}{*}{\ding{55}}
    & \multirow{2}{*}{\ding{55}}
    & \multirow{2}{*}{\ding{55}}
    & 41.11 & 4.44 & 33.33 & 23.89 & \textemdash & 34.22 & 20.89 & 19.11 & 17.33 & \textemdash \\
    & Phi3-V-3.8B
    &  &  & 
    & 46.67 & 9.44 & 15.00 & 20.00 & \textemdash & 51.56 & 22.22 & 28.89 & 15.11 & \textemdash \\ \hdashline

    \multirow{2}{*}{\makecell[c]{PeKit}}
    & InternVL3.5-8B
    & \multirow{2}{*}{\ding{51}}
    & \multirow{2}{*}{\ding{55}}
    & \multirow{2}{*}{\ding{55}}
   & 61.11 & 49.44 & 21.11 & 35.00 & \textemdash & 52.89 & 40.89 & 29.78 & 20.00 & \textemdash \\
    & Qwen3.5-9B
   & & & & 68.33 & 56.67 & 35.56 & 48.33 & \textemdash & 56.00 & 33.78 & 19.56 & 10.22 & \textemdash \\ \hdashline

    \multirow{2}{*}{\makecell[c]{R2P}}
    & InternVL3.5-8B
    & \multirow{2}{*}{\ding{51}}
    & \multirow{2}{*}{\ding{55}}
    & \multirow{2}{*}{\ding{55}}
   & 52.78 & 35.56 & 8.89 & 25.00 & \textemdash & 65.78 & 40.89 & 10.22 & 26.22 & \textemdash \\
    & Qwen3.5-9B
   & & & & 66.67 & 55.56 & 45.00 & 48.33 & \textemdash & 60.89 & 41.78 & 20.00 & 36.44 & \textemdash \\ \hdashline
    
    \multirow{2}{*}{\makecell[c]{TAME$^{\dagger}$}}
    & InternVL3.5-8B
    & \multirow{2}{*}{\ding{51}}
    & \multirow{2}{*}{\ding{51}}
    & \multirow{2}{*}{\ding{51}}
    &  53.89 & 39.44 & 22.78 & 42.78 & 33.08 & 58.22 & 19.11 & 34.67 & 15.11 & 28.73\\
    & Qwen3.5-9B
    &  &  & 
    & 65.56 & 34.44 & 22.22 & 42.22 & 34.59 & 49.33 & 32.44 & 50.67 & 27.11 & 31.63 \\ \hdashline
    
    \multirow{2}{*}{\makecell[c]{\textbf{\M}}}
    & InternVL3.5-8B
    & \multirow{2}{*}{\ding{51}}
    & \multirow{2}{*}{\ding{51}}
    & \multirow{2}{*}{\ding{51}}
        & \cellcolor{gray!10} 75.00 & \cellcolor{gray!10} \textbf{61.67} & \cellcolor{gray!10} 51.11 & \cellcolor{gray!10} 57.78 & \cellcolor{gray!10} 69.17 & \cellcolor{gray!10} 70.67 & \cellcolor{gray!10} 38.22 & \cellcolor{gray!10} 56.44 & \cellcolor{gray!10} 51.11 & \cellcolor{gray!10} 66.37  \\
    & Qwen3.5-9B
    &  &  & 
    & \cellcolor{gray!10} \textbf{82.22} & \cellcolor{gray!10} \textbf{61.67} & \cellcolor{gray!10} \textbf{64.44} & \cellcolor{gray!10} \textbf{70.00} & \cellcolor{gray!10} \textbf{72.18} & \cellcolor{gray!10} \textbf{75.11} & \cellcolor{gray!10} \textbf{52.89} & \cellcolor{gray!10} \textbf{72.00} & \cellcolor{gray!10} \textbf{59.11} & \cellcolor{gray!10} \textbf{71.49} \\ 
    \shline
    \end{tabular}
    }
    \vspace{-3mm}
\end{table*}

%% file: sec/5-Conclusion.tex
\section{Conclusion}
\label{sec:conclusion}
In this study, we pioneered the research in long-context MLLM personalization into the open-set setting.
To this end, we proposed \M, which automatically tames novel personalized concepts from everyday interactions while resolving potential conflicts with previously memorized ones.
Extensive experiments in the open-set setting on LCMP-H and LCMP-O benchmarks demonstrate that \M effectively integrates incoming personalized concepts even under visually similar concepts, mitigates concept fission, and remains robust to varying concept arrival orders.

%% file: sec/7-Appendix.tex
\newpage
\appendix

\crefalias{section}{appendix}

\onecolumn

\section*{Appendix}

\startcontents[appendix]
\section*{Content}
\printcontents[appendix]{}{1}{\setcounter{tocdepth}{2}}
\bigskip

\clearpage

\begin{algorithm}[!t]
\caption{Python-like Pseudocode for \M}
\label{alg:tameo}
\definecolor{codeblue}{rgb}{0.25,0.5,0.5}
\definecolor{codekw}{rgb}{0.85, 0.18, 0.50}
\lstset{
  backgroundcolor=\color{white},
  basicstyle=\fontsize{10pt}{10pt}\ttfamily\selectfont,
  columns=fullflexible,
  breaklines=true,
  captionpos=b,
  commentstyle=\fontsize{7.5pt}{7.5pt}\color{codeblue},
  keywordstyle=\fontsize{7.5pt}{7.5pt}\color{codekw},
  escapeinside={(*@}{@*)}
}
\begin{lstlisting}[language=python]
# Q_i: dialogue at time step i (q_i, a_i)
# q_i: current user query (before generation)
# M_T: Time-Aware Dynamic Memory (Cid -> [(Attr, t)])
# C: personalized concept set
# K: number of candidate concepts
# W: context window size
# T: fusion interval

# initialize memory and concept set
M_T = {}
C = set()

for i, Q_i in enumerate(dialogue_stream):

    # -------- Encode user query --------
    T_i, V_i = Encode(q_i)

    # -------- Concept Reconcile (CR Skill) --------
    scores = {}
    for c_k in C:
        # temporally weighted retrieval using timestamps t_k^(j), each item is (Attr, t)
        scores[c_k] = TemporalWeightedSimilarity(T_i, V_i, M_T[c_k])

    # select Top-K candidate concepts
    C_K = TopK(scores, K)

    # MLLM-based concept reconciliation (Eq. 5)
    Cid* = MLLM_Reconcile(recent_memory(C_K, M_T), q_i, recent_dialogues(W))

    # decide whether concept is novel or existing
    if Cid* == "novel":
        Cid = NewConceptID()
        C.add(Cid)
        M_T[Cid] = []          # initialize memory list
    else:
        Cid = Cid*

    # -------- Personalized Response Generation --------
    # retrieve relevant memory items using resolved Cid
    context = RetrieveMemory(M_T, Cid)
    a_i = MLLM_Generate(q_i, context)

    # -------- Update Time-Aware Dynamic Memory --------
    # delta indicates add / modify of an attribute
    Cid, Attr, delta = MLLM_Update(Cid, Q_i, M_T)

    # update temporal marker for this concept
    t_new = GetNextTimestamp(M_T[Cid])
    M_T = MemoryUpdate(M_T, Cid, Attr, delta, t_new)

    # -------- Periodic Concept Fusion (PCF Strategy) --------
    if i % T == 0:
        # compute pairwise concept similarity
        concept_pairs = PairwiseSimilarity(M_T)
        fusion_pairs = TopL(concept_pairs)

        for (c_m, c_n) in fusion_pairs:
            M_T = MLLM_Fuse(c_m, c_n, M_T, \tau)

\end{lstlisting}
\end{algorithm}

\section{Pseudocode of \M}
\label{app:pseudocode}
A concise version of Python-style pseudocode for \M is offered in~\Cref{alg:tameo}.

\section{Complexity Analysis of \M}
\label{app:complexity}
We delve into the complexity analysis of \M.
We first provide a definition of the complexity of each MLLM invocation:
\begin{definition}[\textbf{Complexity of an MLLM Invocation}]
\label{def:attention}
We consider a single invocation of an MLLM consisting of $H$ multimodal Transformer layers, each with hidden dimension $d$, operating on an input sequence $X$ with length $l$,
the self-attention operation is defined as:
\begin{equation}
\mathrm{Attention}(X) = \mathrm{softmax}\!\left(\frac{QK^\top}{\sqrt{d}}\right)V,
\end{equation}
where $Q = XW_Q$, $K = XW_K$, and $V = XW_V$, with $W_Q, W_K, W_V \in \mathbb{R}^{d \times d}$.
For simplicity, we assume standard multi-head self-attention with shared sequence length for queries, keys, and values.
Under this setting, the computational complexity of one MLLM invocation is:
$
O\big(H \cdot (l d^2 + l^2 d)\big),
$
where $l d^2$ corresponds to the linear projections for query, key, and value representations, and $l^2 d$ arises from the self-attention computation.
However, $H$ is often a small constant, and we therefore simplify the complexity to $
O(l d^2 + l^2 d)
$
in the following discussion.
\end{definition}
We then analyze the computational complexity of \M by decomposing it into two major components:
(1) the Concept Reconciliation Skill (CR Skill), and
(2) the Periodic Concept Fusion Strategy (PCF Strategy).

\subsection{Complexity of CR Skill}

The CR Skill consists of two stages: temporally weighted candidate retrieval and MLLM-based concept resolution.

\textbf{Temporally Weighted Retrieval}.
At time step $i$, the assistant maintains a dynamic concept set $\mathcal{C}_{i-1}$.
Each concept $c_k$ is associated with $N_k$ memory items stored in the TAD Memory.
For an incoming query $q_i$, the CR Skill computes the temporally weighted similarity score $S(c_k \mid q_i)$ by comparing the query representation with all memory items of $c_k$.
The cost of computing a single cosine similarity in the CLIP embedding space is $O(d)$.
Therefore, the total cost of temporally weighted retrieval over all concepts is:
$O\!\left(
\sum_{k=1}^{|\mathcal{C}_{i-1}|} N_k \cdot d
\right)
=
O\!\left(
|\mathcal{M}_T| \cdot d
\right)$,
where $|\mathcal{M}_T|$ denotes the total number of memory items in the TAD Memory.
In practice, $N_k$ is bounded to a constant, making this step linear in memory size.

\textbf{MLLM-based Concept Reconciliation}.
After retrieval, only the Top-K candidate concepts are retained, where $K \ll |\mathcal{C}_{i-1}|$.
The CR Skill then jointly feeds the most recent memory items of these candidate concepts, together with the current user query and a fixed window of recent dialogue turns, into the MLLM for final concept resolution.
Importantly, this reconciliation step is performed via a single MLLM invocation, rather than multiple independent calls per candidate.
Since the number of candidate concepts $K$, the number of memory items per candidate, and the dialogue context window size are all strictly bounded, the total input sequence length to the MLLM is also bounded.
Consequently, the MLLM-based concept resolution involves a bounded-length attention computation.
According to the attention complexity, this step incurs a cost of
$ O(l d^2 + l^2d),
$
where $l$ is the total token length of the fixed prompt, candidate memory items, and recent dialogue context.
As $l$ is independent of the total number of discovered concepts $|\mathcal{C}_{i-1}|$, this step introduces a bounded computational overhead.

\subsection{Complexity of PCF Strategy}

The PCF Strategy is executed periodically every $T$ interaction steps, rather than at every step.
At each execution, the strategy computes pairwise similarities between personalized concepts to identify potentially forked ones.

For each concept pair $(c_m, c_n)$, the similarity $S(c_m, c_n)$ is computed by averaging cosine similarities over all pairs of memory items from $\mathcal{M}_m$ and $\mathcal{M}_n$.
The cost of computing the similarity for one concept pair is:
$O\!\left(
|\mathcal{M}_m| \cdot |\mathcal{M}_n| \cdot d
\right)$.
Subsequently, computing similarities for all concept pairs yields:
$O\!\left(
\sum_{(c_m, c_n) \in \mathcal{C}\times\mathcal{C},\, m\neq n}
|\mathcal{M}_m| \cdot |\mathcal{M}_n| \cdot d
\right)$.
For each selected pair, the actual fusion decision is delegated to the MLLM.
Similar to the CR Skill, this involves a bounded-length attention computation and incurs a constant overhead.
Moreover, since we restrict the number of candidate pairs for concept fusion to the Top-$L$ most similar ones that also exceed a predefined similarity threshold, the overall computational complexity of this step is strictly bounded.
\textbf{In practice}, we observe that \M incurs about 6.5 seconds runtime cost per user-MLLM interaction, which is acceptable.
And baseline TAME$^{\dagger}$ requires about 5 seconds per interaction.

\begin{figure*}[t]
\centering
\includegraphics[width=0.95\linewidth]{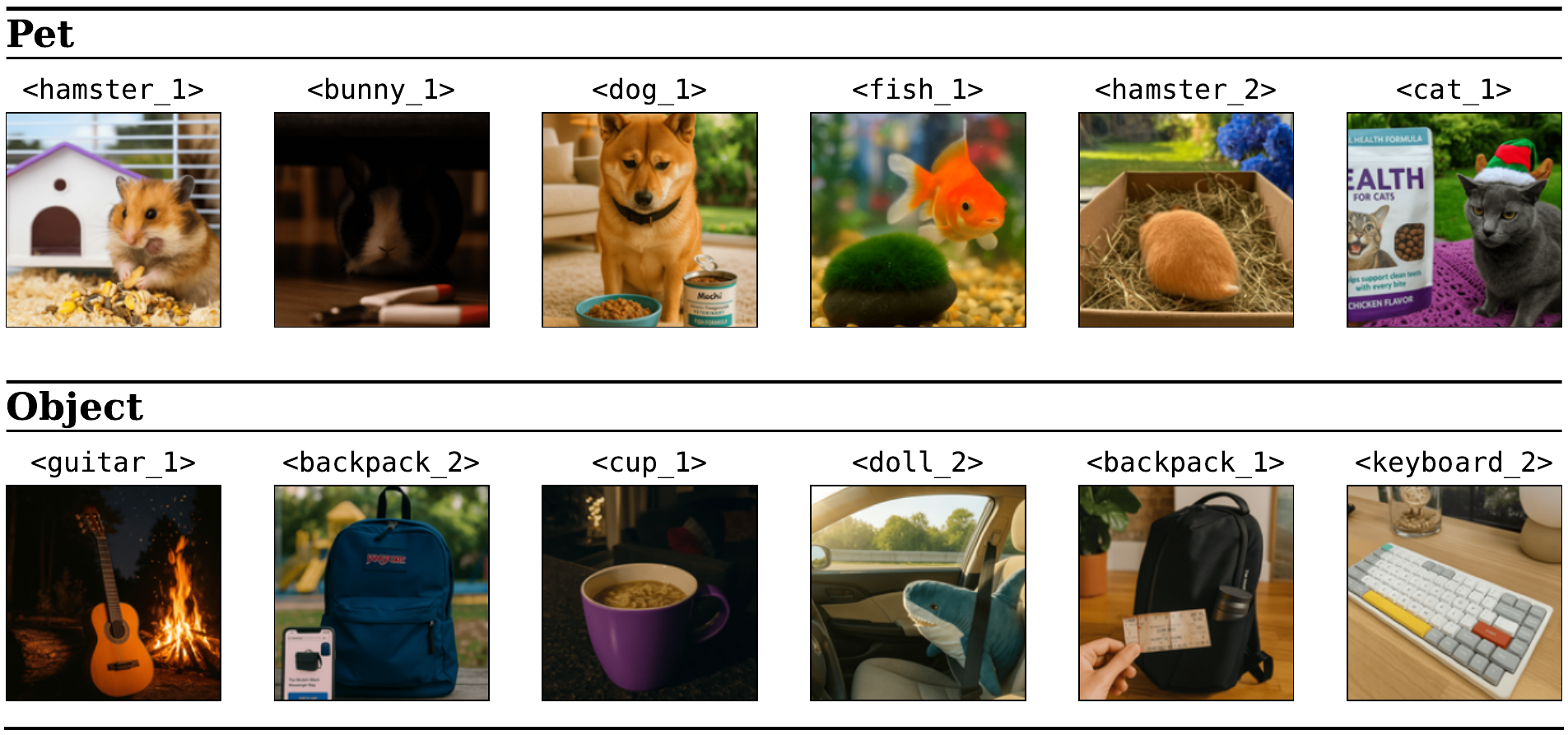}
\vspace{-2mm}
\caption{Cases of personalized concepts from LCMP benchmark.
}
\label{fig-app:lcmp-concept-case}
\vspace{-3mm}
\end{figure*}

\section{Detailed Experimental Settings}
\label{app:experimental_setup}

\subsection{Benchmark: LCMP-H}
Since open-set personalization is studied in long-context, real-world human–MLLM interactions, rather than in static and overly simplified settings (e.g., personalized image captioning), we first adopt the hard version of the LCMP benchmark (Long-Context MLLM Personalization)~\cite{hong2026tameing}, a recent benchmark and the only available benchmark specifically designed for long-context MLLM personalization.
Below we propose a detailed description of this benchmark.

LCMP is the first benchmark for evaluating personalized multimodal large language models in long-context interactions. 
It consists of 30 personalized concepts, including Pet (10), Object (10), and Person (10), along with approximately 420 historical dialogue turns, 240 easy questions, 180 hard questions, and 720 personalized images, and includes two evaluation sets:
(1) \textbf{\B-E}, which contains easy VQA questions involving a single type of attribute (either short-term or long-term), and
(2) \textbf{\B-H}, which contains hard questions requiring reasoning over both types of attributes.
It measures an assistant's ability to perceive and model variations in both short-term and long-term attributes of user-specific concepts and to generate contextually appropriate responses based on extended dialogue histories.
LCMP is constructed through three stages: 
(1) personalized concept creation, where raw images are sampled and endowed with fine-grained temporal attributes; 
(2) multi-turn vision–language dialogue generation, which simulates historical interactions that reveal concept attribute changes over time; 
and (3) evaluation question generation, which produces single-turn VQA tasks requiring reasoning over the preceding dialogue context. 
The benchmark thus explicitly tests long-context reasoning and personalization beyond static single-turn tasks.
We adopt the hard version of the LCMP benchmark, namely LCMP-H, in this study for evaluation.
As illustrated in~\Cref{fig-app:lcmp-concept-case}, we provide cases of personalized concepts from LCMP-H.

\subsection{Benchmark: LCMP-O}
\label{app:lcmp-o}
To comprehensively evaluate the open-set MLLM personalization capability of the MLLM assistant, we construct a new benchmark: \textbf{O}pen-Set \textbf{L}ong-\textbf{C}ontext \textbf{M}LLM \textbf{P}ersonalization (\textbf{LCMP-O}).
The original LCMP benchmark exhibits two limitations in evaluating open-set personalization performance:
\textbf{(1)} LCMP contains a small number of personalized concepts, which constrains the diversity of the open-set evaluation.
\textbf{(2)} The personalized concepts in LCMP are characterized by highly distinctive attributes, making it relatively easy for MLLMs to differentiate new concepts from previously grasped ones, softening the challenges of open-set personalization.

Taking the above limitations into consideration, we construct LCMP-O with 45 personalized concepts, deliberately designed to present high intra-class visual similarity and thus pose stronger challenges for open-set personalization evaluation.
Following the construction pipeline of LCMP~\cite{hong2026tameing}, we generate 10 historical dialogue turns and 5 hard evaluation questions per concept, yielding a total of 450 dialogues and 225 evaluation questions.

The 45 concepts span three types, each chosen to maximize within-type visual ambiguity:
(1) 15 Pets (cats, dogs, and rabbits), where every three concepts belong to the same breed, making fine-grained visual disambiguation particularly challenging. These cats and dogs concepts are sourced from the Oxford-IIIT-Pet dataset~\cite{parkhi2012cats}, which covers 37 breeds with substantial intra-breed variation in pose, lighting, and deformation. And the rabbits concepts are collected by authors.
(2) 15 Persons, sampled from subjects of similar age with an approximately 1:1 gender ratio. Person images are drawn from MS-COCO~\cite{lin2014microsoft}, following the same protocol as TAME~\cite{hong2026tameing}.
(3) 15 Objects, where every three concepts belong to the same object class. 
These concepts are sourced from PerVA~\cite{seifi2025personalization}, which contains object concepts that exhibit high visual and semantic similarity.
In~\Cref{fig-app:lcmp-o-concept-case}, we provide cases of concepts from the LCMP-O benchmark.

\subsection{Baselines}
\label{app:baselines}
Since open-set MLLM personalization is novel as a setting, we compare \M with traditional closed-set MLLM personalization baselines and a specifically designed baseline for the open-set setting.
For traditional baselines, we select MyVLM~\cite{alaluf2024myvlm}, Yo'LLaVA~\cite{nguyen2024yollava}, RAP~\cite{hao2024rap}, PeKit~\cite{seifi2025personalization}, and R2P~\cite{das2025training}, with detailed descriptions as follows:
\begin{itemize} [leftmargin=*, topsep=2pt, partopsep=0pt, itemsep=0pt]

\item \textbf{MyVLM} pioneers the personalization of general VLMs so they can understand and talk about concepts unique to a given user. 
It augments LLaVA~\cite{liu2023visual} with external concept heads that first detect a target entity and then inject a learned concept embedding into the model's mid-level features, guiding the decoder to mention the concept naturally.

\item \textbf{Yo'LLaVA} frames subject personalization as learning a compact representation from only a few example images.  
It encodes the subject into a small set of latent prefix tokens that sit between the vision encoder and LLaVA's language decoder, binding visual attributes to text without touching backbone weights.

\item \textbf{RAP} turns a generic MLLM into a user-aware assistant by adding an external memory layer. A key–value store ``remembers'' user facts and portraits, a multimodal retriever selects the most relevant keys at inference, and the generator conditions on both the query and retrieved context to tailor its answer.

\item \textbf{PeKit} argues that strong personalization can be achieved entirely at inference time. It pairs off-the-shelf visual descriptors with retrieval-augmented generation and lightweight visual prompts, injecting the correct identity or object instance into any MLLM decoder without gradient updates.

\item \textbf{R2P} shows that an MLLM can personalize itself using nothing but its pre-trained weights and a clever retrieval stack. It distills each personalized concept into a sparse fingerprint of discriminative attributes, retrieves candidate fingerprints at run time, and employs chain-of-thought reasoning to verify them before generation.

\end{itemize}

For open-set personalization method, we additionally create a competitive baseline \textbf{TAME$^{\dagger}$}, a modified variant of the long-context MLLM personalized assistant TAME~\cite{hong2026tameing}.
TAME is the first long-context MLLM personalization method that explicitly models both the long-term and short-term variations of personalized concepts and retrieves reliable contextual information for each concept to support personalized response generation.
We unlock the open-set capability of TAME by retrieving the most relevant concept stored in memory based on the visual and textual information in the query.
A predefined similarity threshold is then applied: if the retrieved similarity exceeds the threshold, the concept is classified as existing; otherwise, it is treated as novel.
We conduct a search over the threshold in the range $[0, 1]$ with an interval of $0.1$, and select the best-performing value of $0.6$ for all experiments with TAME$^{\dagger}$.

\subsection{Implementation Details}
\label{app:implementation-details}
We adopt the latest released MLLMs \texttt{Qwen3.5-9B}~\cite{qwen3.5}, and \texttt{InternVL3.5-8B}~\cite{wang2025internvl}, as the MLLM backbone.
Since \M is a plug-and-play framework, we adopt TAME as the base personalized assistant for evaluation.
When identifying the concept in user's query, we first prompt the MLLM to extract open-set category descriptions from the query, and then use the extracted categories to perform grounding on the query image with \texttt{Grounding-DINO}~\cite{liu2024grounding}. 
In addition, we adopt the \texttt{Qwen3-VL-Embedding-2B}~\cite{li2026qwen3vlembedding} to generate embeddings for both visual and textual inputs.  
For the CR Skill, we set the temporal decay coefficient $\lambda$ to 1, the number of candidates $K$ to 3, and the number of latest memory items for each concept $R$ to 3.
For PCF Strategy, we set the time interval $T$ to 10, the number of fusion candidates $L$ to $3$, and the threshold $\tau$ to $0.6$.
For the CFR metric in Eq.~(\ref{eq:10}), we set $\epsilon$ to 1.
Since each forked identifier is created at one of the $t - |\mathcal{C}_t|$ turns that revisit an already-seen concept, $\mathrm{CFR}(t) \in [0, 1]$.
All the experiments are conducted on an NVIDIA RTX 5090 GPU.

\textbf{Concept Identifiers.}
In our experiments, each novel concept is assigned an identifier named after its ground-truth concept in the benchmark (e.g., \texttt{dog\_2}), and each forked concept receives an additional numeric suffix (e.g., \texttt{dog\_2\_1}).
These identifiers serve only as evaluation bookkeeping, allowing CRA and CFR to map every predicted identifier back to its ground-truth concept.
They are never shown to the MLLM, since an identifier is assigned only after the CR Skill has decided that a concept is novel, and both $\mathcal{P}_r$ and $\mathcal{P}_f$ refer to candidate concepts through anonymous labels, describing each one only by its memory items and portrait (cf.~\Cref{app:prompt}).
Therefore, the identifiers leak no ground-truth information, and because all compared methods, including TAME$^{\dagger}$, share the same scheme, none of them benefits from it.

\section{Hyper-Parameter Analysis}
\label{app:hyper}
In this section, we evaluate the effect of important hyper-parameters within \M: the temporal decay coefficient $\lambda$, and the number of candidates $K$.
As shown in~\Cref{tab-app:hyper}, our \M exhibits relatively low sensitivity to hyperparameter variations, when $\lambda > 0$ or $K > 1$.
We set $K = 3$ and $\lambda = 1$ across all datasets.

\begin{table*}[!t]
\centering
\caption{Performance of \M under varying values of hyper-parameters.}
\label{tab-app:hyper}
\setlength{\tabcolsep}{9pt}
   \resizebox{\linewidth}{!}{
\begin{tabular}{ccccccccccc}
\toprule

& \multicolumn{5}{c}{Parameter $\lambda$} 
& \multicolumn{5}{c}{Parameter $K$} \\
\cmidrule(lr){2-6} \cmidrule(lr){7-11}
\textbf{LCMP-H} & 0 & 0.1 & 1 & 2 & 5 
& 1 & 2 & 3 & 4 & 5 \\
\midrule
ACC-C (\%) 
& 78.33 & 80.56 & 82.22 & 82.44 & 78.67
& 74.44 & 82.78 & 82.22 & 81.11 & 80.00 \\
\bottomrule

& \multicolumn{5}{c}{Parameter $\lambda$} 
& \multicolumn{5}{c}{Parameter $K$} \\
\cmidrule(lr){2-6} \cmidrule(lr){7-11}
\textbf{LCMP-O} & 0 & 0.1 & 1 & 2 & 5 
& 1 & 2 & 3 & 4 & 5 \\
\midrule
ACC-C (\%) 
& 72.89 & 75.89 & 75.11 & 75.44 & 76.11
& 71.44 & 74.17 & 75.11 & 75.67 & 75.18 \\
\bottomrule
\end{tabular}}
\end{table*}

\begin{table*}[!t]
\centering
\caption{CFA of \M over successive time steps on different datasets.}
\label{tab-app:cfa}
\resizebox{\linewidth}{!}{
\setlength{\tabcolsep}{9.5pt}
\begin{tabular}{ccccccccccc}
\toprule
& \multicolumn{5}{c}{\textbf{LCMP-H}} 
& \multicolumn{5}{c}{\textbf{LCMP-O}} \\
\cmidrule(lr){2-6}
\cmidrule(lr){7-11}
\textbf{Time Step (t)} & 50 & 100 & 200 & 300 & 400
& 50 & 100 & 200 & 300 & 400 \\
\midrule
CFA (\%)
& 100.00 & 100.00 & 100.00 & 100.00 & 100.00
& 100.00 & 100.00 & 94.74 & 95.65 & 96.67 \\
\shline
\end{tabular}}
\end{table*}

\section{Quantitative Analysis of the PCF Strategy}
\label{app:pcf-strategy}
In the main paper, we offer the Concept Fission Rate (CFR) metric for \M when equipped with the PCF Strategy.
Here, we take a further step to empirically evaluate the fusion accuracy of the PCF Strategy.
Formally, each fusion operation is defined as an automatic merging process that combines two personalized concepts $c_m$ and $c_n$ stored in the TAD Memory into a unified one.
When $c_m$ and $c_n$ correspond to the same ground-truth concept, the fusion operation is considered correct; otherwise, it is regarded as incorrect.
Based on this criterion, we define the Concept Fusion Accuracy (\textbf{CFA}) as follows.
\begin{equation}
\mathrm{CFA}(t)
=
\frac{1}{|\mathcal{U}_{\le t}|}
\sum_{(c_m, c_n) \in \mathcal{U}_{\le t}}
\mathbb{I}\!\left(
\texttt{GT}(c_m) = \texttt{GT}(c_n)
\right),
\end{equation}
where $\mathcal{U}_{\le t}$ denotes the set of concept pairs $(c_m, c_n)$ that have been accepted by the PCF Strategy and fused into unified concepts up to time step $t$,
$\texttt{GT}(c_m)$ denotes the ground-truth concept identifier of concept $c_m$, which corresponds to the identifier assigned when the concept first appears,
and $\mathbb{I}(\cdot)$ is a binary indicator function that outputs 1 if the condition holds and 0 otherwise.
In~\Cref{tab-app:cfa}, we present the cumulative CFA at successive time steps while processing the historical dialogues in LCMP-H and LCMP-O.
We observe that \M consistently achieves high CFA values, suggesting that the fusion decisions made by the PCF Strategy are reliable.

\textbf{Note on the CFR Computation.}
All CFR values in~\Cref{fig:concept-fission} are computed strictly following Eq.~(\ref{eq:10}) with $\epsilon = 1$, recomputed from the logs of the same runs, and plotted without any per-method rescaling.
This corrects two inconsistencies in the submitted version.
First, the submitted Figure~6 normalized the total number of identifiers $\sum_{c_i \in \mathcal{C}_t} |\hat{\mathcal{I}}_{c_i}^{(t)}|$, rather than the number of forked identifiers, by $t - |\mathcal{C}_t|$, and rescaled the TAME$^{\dagger}$ curve by a factor of 0.55.
Second, the logger in the released code normalized the number of forked identifiers by the total number of identifiers instead of by $t - |\mathcal{C}_t|$.
Both the plotting script and the released code are now aligned with Eq.~(\ref{eq:10}).
Under all three normalizations, the ordering TAME$^{\dagger}$ $>$ \M w/o PCF Strategy $>$ \M holds at every sampled time step.

\begin{table*}[t]
    \caption{Performance comparison on the LCMP-H and LCMP-O datasets.
    Higher values indicate better performance.}
    \vspace{-1mm}
    \label{tab-app:scalability}
    \centering
    \renewcommand{\arraystretch}{1.1}
    \setlength{\tabcolsep}{4.5pt}
    \resizebox{\linewidth}{!}{
    \begin{tabular}{l|c|ccccc|ccccc}
    \shline
     \multirow{2}{*}{\textbf{Methods}} 
     & \multirow{2}{*}{\textbf{Open-Set Method}} 
     & \multicolumn{5}{c|}{\textbf{LCMP-H}} 
     & \multicolumn{5}{c}{\textbf{LCMP-O}} \\ 
     & 
     & \textbf{ACC-C} & \textbf{ACC-F} & \textbf{SPR-L} & \textbf{SPR-S} & \textbf{CRA}
     & \textbf{ACC-C} & \textbf{ACC-F} & \textbf{SPR-L} & \textbf{SPR-S} & \textbf{CRA} \\
    \hline 

    \multirow{2}{*}{\makecell[c]{PeKit}}
    & TAME$^{\dagger}$
    & 41.67 & 26.67 & 18.33 & 23.33 & 24.56
    & 38.22 & 22.22 & 16.00 & 10.22 & 19.82 \\
    & \textbf{\M}
    & \textbf{48.33} & \textbf{45.56} & \textbf{28.89} & \textbf{37.22} & \textbf{49.62}
    & \textbf{48.00} & \textbf{26.22} & \textbf{20.00} & \textbf{12.44} & \textbf{48.33} \\
    \hdashline

    \multirow{2}{*}{\makecell[c]{R2P}}
    & TAME$^{\dagger}$
    & 36.11 & 26.11 & 26.11 & 23.33 & 25.56
    & 32.44 & 21.33 & 16.00 & 28.44 & 23.16 \\
    & \textbf{\M}
    & \textbf{42.22} & \textbf{31.67} & \textbf{36.67} & \textbf{35.00} & \textbf{54.14}
    & \textbf{50.22} & \textbf{34.22} & \textbf{17.33} & \textbf{30.22} & \textbf{52.34} \\ 
    \bottomrule
    \end{tabular}}
    \vspace{-2mm}
\end{table*}

\section{Scalability across Other MLLM Personalized Assistants}
\label{app:scalability}
Since \M is designed as a training-free and plug-and-play framework, we further evaluate its scalability by integrating it with other personalized MLLM assistants beyond TAME~\cite{hong2026tameing} (reported in the main paper).
Specifically, we integrate \M with two representative training-free, RAG-based personalization methods, PeKit\cite{seifi2025personalization} and R2P\cite{das2025training}, which differ in their memory organization and retrieval strategies. 
To enable open-set personalization, we augment both methods with our Concept Reconciliation (CR) Skill and a Time-Aware Dynamic (TAD) Memory that stores dialogue contexts with temporal markers. 
The CR Skill is used to determine whether an incoming personalized concept is novel or previously observed, while the original memory mechanisms of each method are retained for concept-specific storage and retrieval during response generation.
As illustrated in~\Cref{tab-app:scalability}, we compare the performance of these two baseline methods under the open-set setting by integrating TAME$^{\dagger}$ and \M, respectively.
We observe that integrating \M yields superior open-set personalization performance, demonstrating the strong scalability and generality of our method.

\section{Additional Literature Review}
\label{app:additional-literature-review}

\subsection{Memory-Augmented LLMs}
Recent research on LLMs explores augmenting models with external memory to store and retrieve past interactions, enabling improved adaptability without parameter updates~\cite{zhao2024expel,zhang2025survey,liu2026exploratory}. 
Existing studies typically organize memory as structured summaries, temporal events, or reasoning traces to highlight key information while reducing redundancy~\cite{maharana2024evaluating,liu2023thinkinmemory}. 
To scale memory usage, retrieval mechanisms are introduced to efficiently locate relevant past experiences, often embedding memory entries as dense vectors and retrieving the most relevant ones via similarity search~\cite{zhong2024memorybank,penha2024bridging,xu2025memoryaugmented}, with cosine similarity and vector indexing tools such as FAISS commonly used in practice. 
These studies demonstrate that external memory can effectively extend LLM capabilities without retraining by enabling models to accumulate and reuse past knowledge.

Building upon these advances, memory-based methods have also been studied in MLLM personalization, eliminating the need for frequent retraining for novel concepts~\cite{hao2024rap,hong2026tameing}. 
Unlike general memory-augmented LLM approaches that treat each memory entry independently, these methods explicitly maintain concept-level memory units to organize information around individual personalized concepts.
Approaches such as RAP~\cite{hao2024rap} and PeKit~\cite{seifi2025personalization} store user-specific concepts in external memory and perform retrieval-based matching, while R2P~\cite{das2025training} introduces structured ``fingerprints'' with reasoning for identification. 
More recently, TAME~\cite{hong2026tameing} extends the memory for concept storage to the modeling of long-context information for each user-specific concept. 
However, these methods generally assume a closed-set of predefined concepts and fail to extend the memory autonomously by the MLLMs for accommodating novel concepts.
In contrast, our TAME-O introduces a time-aware dynamic memory with explicit concept reconciliation, enabling continuous concept acquisition and refinement under open-set, long-context interactions.
Moreover, the PCF strategy further refines the memory to mitigate excessive concept fission that could otherwise lead to severe memory chaos.

\subsection{Open-Set Recognition}
Open-set recognition (OSR) addresses the limitation of the closed-set assumption in traditional deep learning systems, enabling models to both classify known classes and detect inputs belonging to previously unseen classes~\cite{wang2026classifierinduced,fu2025reason,scheirer2012toward}. 
The problem was first formally introduced by~\cite{scheirer2012toward}, who proposed a binary SVM-based framework for training and evaluation. 
Since then, extensive research has been devoted to this area. 
Existing approaches can be broadly divided into discriminative and generative methods. 
Discriminative approaches aim to learn more robust representations or decision boundaries to reduce overconfidence on unknown classes, such as placeholder-based representative points, supervised contrastive learning, and diverse representation regularization~\cite{zhou2021learning,xu2023contrastive,hang2024binary}. 
In contrast, generative approaches synthesize instances near the boundary between known and unknown classes using techniques such as GANs, variational autoencoders, or prototype-based modeling to better characterize open space risk~\cite{oza2019c2ae,sun2020conditional,guo2021conditional}.

Despite the name ``open-set,'' these methods are primarily designed to identify whether an input belongs to an unknown category, typically labeling it as unknown. 
In contrast, the open-set personalization setting considered in this work requires not only identifying novel user-specific concepts, but also assigning them new personalized concept identities, enabling the assistant to continually expand and refine the set of personalized concepts during interaction.

\subsection{Multimodal Large Language Models}
Large Language Models (LLMs), built upon extensive world knowledge, have rapidly advanced across a wide range of domains~\cite{shen2024taskbench,chang2024survey}. 
Building on this foundation, Multimodal Large Language Models (MLLMs) have recently emerged by integrating language decoders with multimodal encoders, enabling LLMs to perceive and reason over diverse data modalities~\cite{huang2025learn,lian2025affectgpt,jung2025visual,chang2025lvxattn,luo2024feast,kao2024bridging,gu2024mllmguard,liu2024iibench,lang2026shallow}.
Among MLLMs, vision–language models have been extensively studied~\cite{gpt52025openai,wang2025internvl}, and the fusion of visual and linguistic modalities has driven significant progress in applications such as Visual Question Answering (VQA)~\cite{wang20253d}, multimodal medical analysis~\cite{zhu2025mmedpo}, and autonomous driving~\cite{liu2024reasoning}. 
With these advances, MLLMs are increasingly regarded as reliable human assistants, supporting users in various aspects of daily life.
Despite their success, mainstream MLLMs are primarily trained for general-purpose scenarios and fail to adapt to individual users' preferences and personalized contexts~\cite{alaluf2024myvlm}. 
To bridge this gap, recent studies have begun to explore personalized MLLM assistants~\cite{nguyen2024yollava,pi2025personalized,hong2026tameing}.
Among them, MC-LLaVA~\cite{an2024mcllava} extends personalization to multiple concepts via multi-concept instruction tuning, and CaT~\cite{an2026concept} synthesizes training data to strengthen personalized VLMs, while both still operate on a predefined set of concepts.
In this work, we take a step further by addressing the open-set personalization setting and propose the first open-set personalized MLLM assistant \M, which better aligns with users in open-world, long-term interactions.

\subsection{Large Language Model Personalization}
Large Language Model (LLM) personalization~\cite{salemi2024lamp, magister2024personalization, afzoon2024persobench,zhang2018personalizing,jang2022call} focuses on adapting an LLM's responses to a single user's preferences, interaction history, and persona characteristics (e.g., a preference for concise responses).
Early efforts such as PersonaMem~\cite{jiang2025know} introduce benchmarks with numerous simulated profiles to test whether models can leverage dialogue history to infer user preferences and deliver personalized responses. 
PersoBench~\cite{afzoon2024persobench} further explores persona-aware dialogue generation in a zero-shot setting. 
Building upon these evaluations, methods like PLUM~\cite{magister2024personalization} employ dialogue-based data augmentation to fine-tune LLMs for more consistent user-aware behavior. 
PALACE~\cite{liu2025personaaware} advances this line with a topic-aware retrieval mechanism that extracts relevant cues from extended histories and a persona-driven prompt learning module for multi-session adaptation.
Nevertheless, LLM personalization focuses on tailoring a model's responses to a single user's preferences, dialogue history, and persona traits (e.g., preferring concise answers).
In contrast, MLLM personalization emphasizes enabling personalized conversations grounded in user-related multimodal concepts, such as the user's pets, friends, and frequently mentioned objects.

\section{Additional Case Study}
\label{app:case}
\subsection{Additional Cases of Concept Reconciliation}
\begin{figure*}[t]
\centering
\includegraphics[width=\linewidth]{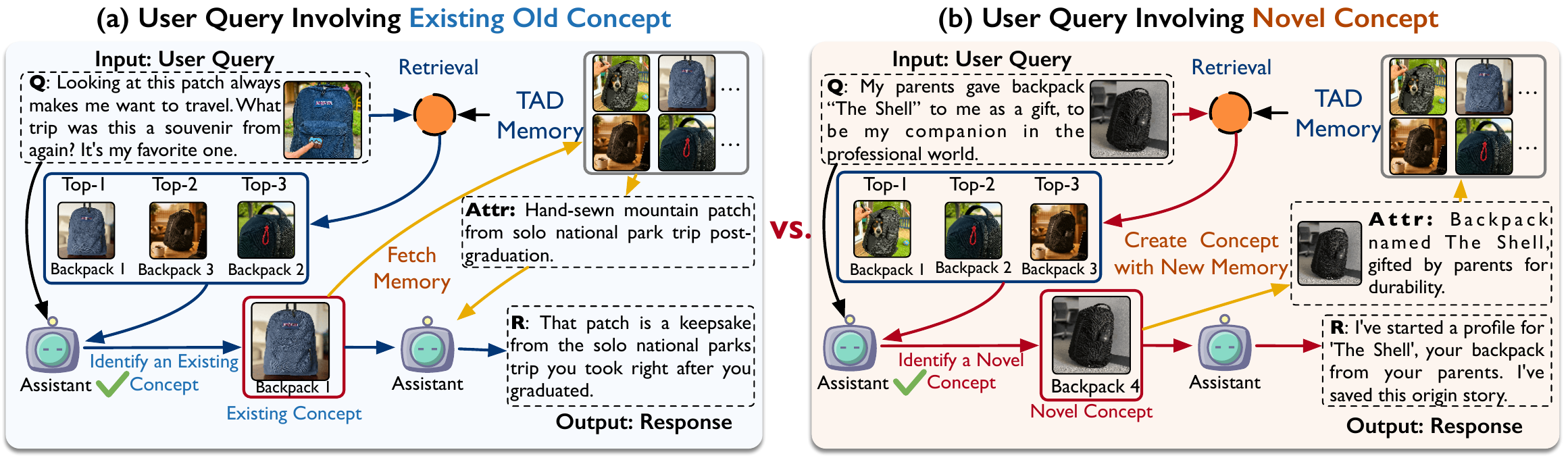}
\vspace{-4mm}
\caption{Case study: the procedure of \M in resolving an existing old concept (left) and a novel concept (right).
}
\label{fig-app:case-study-cr}
\vspace{-4mm}
\end{figure*}

In~\Cref{fig-app:case-study-cr}, we provide additional concept reconciliation cases of \M in resolving novel and existing concepts.

\subsection{Cases of Concept Fusion}
\begin{figure*}[t]
\centering
\includegraphics[width=\linewidth]{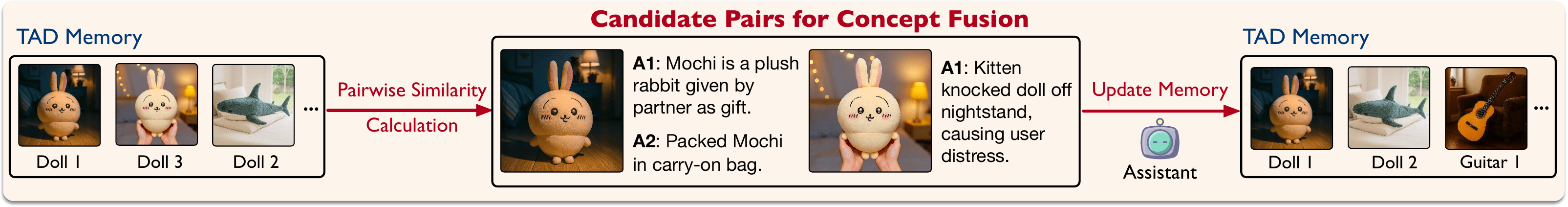}
\vspace{-4mm}
\caption{Case study: two concepts that have similar attributes are selected to be fused into one concept.
}
\label{fig-app:case-study-pcf-strategy}
\vspace{-4mm}
\end{figure*}

As illustrated in~\Cref{fig-app:case-study-pcf-strategy}, we provide a case illustrating the concept fusion procedure of our proposed PCF Strategy.

\begin{figure*}[!t]
\centering
\includegraphics[width=0.6\linewidth]{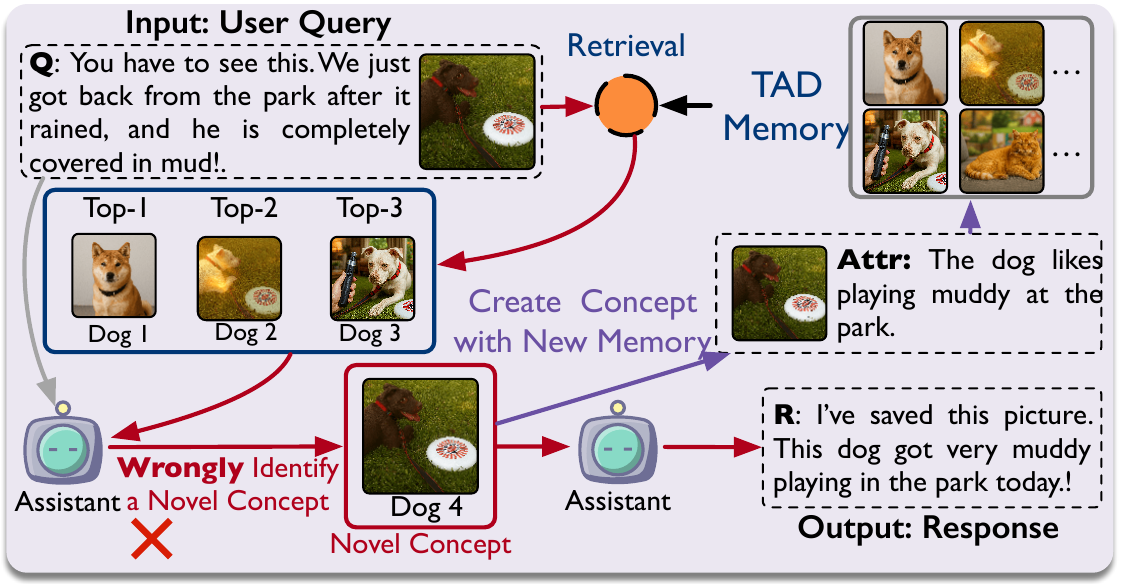}
\vspace{-1mm}
\caption{Case study: incorrect concept identification leads to the fission of an already grasped concept.
}
\label{fig-app:case-study-fission}
\vspace{-4mm}
\end{figure*}

\section{Failure Case Analysis: Concept Fission}
\label{app:failure-analysis}
In this section, we present a failure case analysis of \M, focusing on unintentional concept fission, as illustrated in~\Cref{fig-app:case-study-fission}. Specifically, a concept originally associated with \texttt{Dog 2} --- a pet dog undergoing a temporary physical appearance change (e.g., its fur becoming dirty after playing in the mud) --- is incorrectly identified as a novel concept, \texttt{Dog 4}. This misclassification triggers an unintended concept fission.

\section{Prompt Presentation for \M}
\label{app:prompt}
In this section, we present the prompts $\mathcal{P}_r$ and $\mathcal{P}_f$ designed for performing concept reconciliation and fusion, respectively.

\section{Limitations}
\label{app:limitations}

\textbf{Reliance on External Modules.}
\M relies on off-the-shelf grounding and embedding models to localize and retrieve personalized concepts, and their errors may propagate to concept reconciliation.
Nevertheless, these modules can be readily replaced by stronger ones as they become available.

\textbf{Robustness under Noisy Interactions.}
Our benchmarks are constructed in a controlled manner, whereas real-world interactions can be noisier, e.g., with ambiguous references or low-quality images and descriptions.
We leave a systematic evaluation under such noisy interactions to future work.

\textbf{Indirect Recovery from Incorrect Fusion.}
Concept fusion may occasionally merge two distinct concepts.
Although such a merge cannot be directly reverted, \M can recover from it indirectly: benefiting from the attribute-level and time-aware memory organization, subsequent interactions progressively correct the affected concept memory with recent observations.
A direct mechanism to revert incorrect fusions is left to future work.

\textbf{Scalability and Privacy.}
The pairwise similarity computation in concept fusion grows quadratically with the number of concepts, although the number of MLLM invocations per fusion round is bounded by $L$.
In large-scale deployment, this cost can be reduced by caching pairwise similarities and recomputing only the pairs involving recently updated concepts, or by approximate nearest-neighbor search.
Moreover, long-term personalized memory contains private user information, which calls for local storage and user-controlled deletion in practical deployment.

\section{Broader Impacts}
\label{app:broader-impacts}

Since this study focuses on designing open-set personalized MLLM assistant, which is closely related to everyone's daily life, it has broader impacts in both positive and negative aspects.
\textbf{On the positive side}, open-set MLLM personalization can make future assistants more adaptive, user-aligned, and convenient in daily life. By automatically discovering and maintaining user-specific concepts through long-term interactions, such assistants can provide responses that are more context-aware and personalized, reducing the burden on users to manually register or describe every personal entity. This capability may benefit various real-world scenarios, such as daily assistance, education, accessibility support, elderly care, and long-term human AI collaboration, where understanding user-specific objects, people, pets, or environments is important for providing useful and grounded assistance.
\textbf{On the negative side}, such systems may introduce risks from both misuse and the inherent limitations of MLLMs. For misuse, the ability to continuously acquire personalized concepts could be abused for intrusive profiling, unauthorized surveillance, or manipulative personalized content generation. For model limitations, our method may inherit hallucination and bias issues from the underlying MLLMs. It may generate responses that are not fully grounded in the input image, dialogue context, or stored personalized memories, and such hallucinations can be harmful in high-stakes scenarios. Moreover, biases from the vision encoder and language model may lead to unfair or stereotypical interpretations of user-specific entities. Therefore, open-set personalized MLLM assistants should be deployed with appropriate safeguards, usage restrictions, and careful evaluation, especially in sensitive applications.

\section{License of Assets}
\label{app:license-assets}
\textbf{Datasets}:
LCMP (research-use), Oxford-IIIT Pet (CC BY-SA 4.0), MS-COCO (Flickr Terms of Use and image-specific licenses), and PerVA (research-use).

\input{sec/prompt}

\begin{figure*}[p]
\centering
\includegraphics[width=0.95\linewidth]{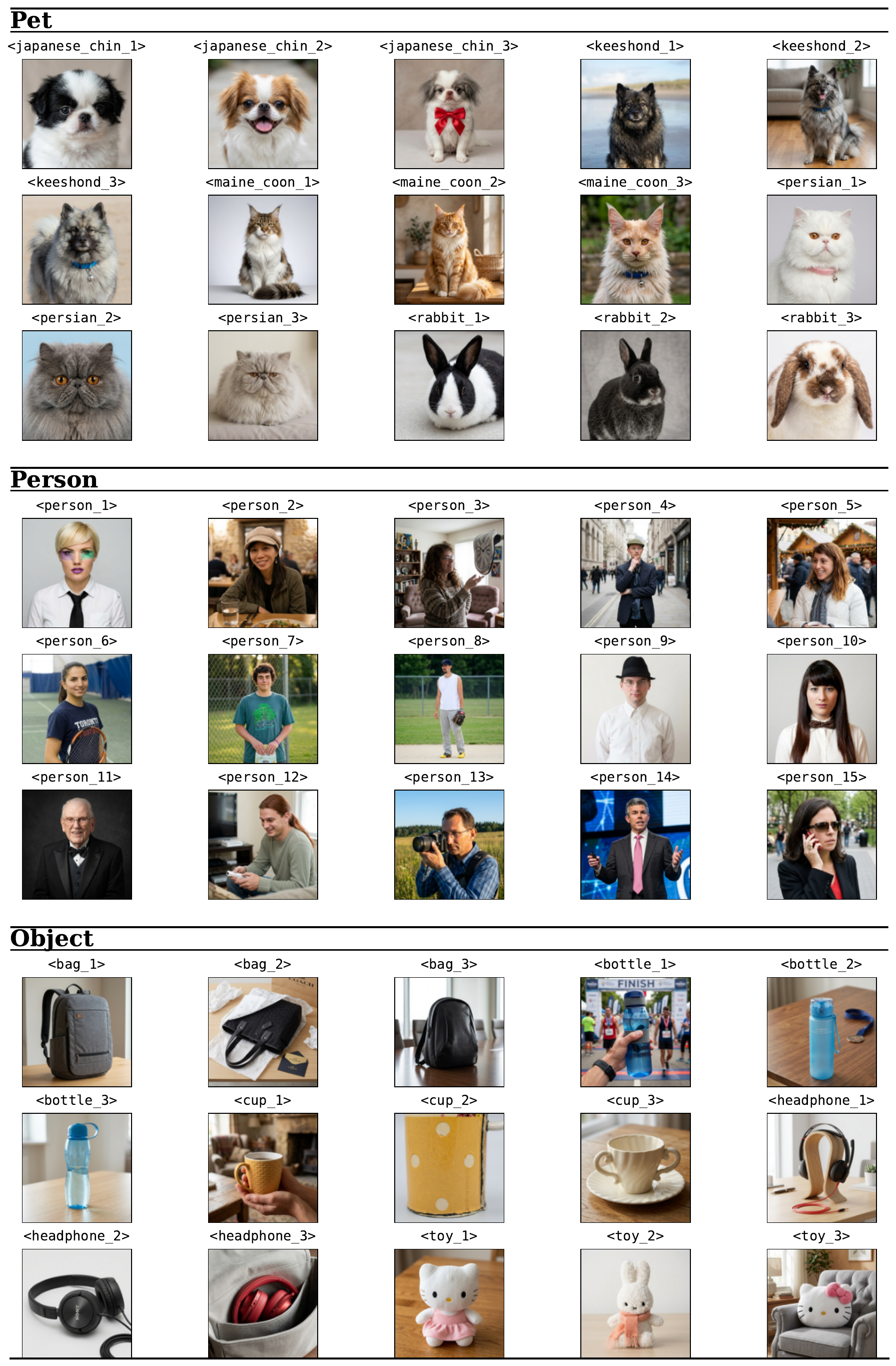}
\vspace{-2mm}
\caption{Cases of personalized concepts from LCMP-O benchmark.
}
\label{fig-app:lcmp-o-concept-case}
\vspace{-3mm}
\end{figure*}

%% file: sec/prompt.tex
\begin{figure*}[!t]
\centering

\begin{tcolorbox}[
  colback=gray!10,
  colframe=black,
  boxrule=0.1mm,
  width=\textwidth,
  sharp corners,
  title=\textbf{Prompt $\mathcal{P}_r$ for Concept Reconciliation.}
]
\label{app:prompt-pr}

\textbf{Prompt:}
You are a concept identification assistant. Your goal is to determine if the user's question refers to one of the EXISTING candidates or to a completely NEW concept. \\
\{\oc{\textbf{Recent Conversations}}\}\\
\vspace{0.12em}
\textbf{\# Question}\\
\{\oc{\textbf{User Query}}\}\\
Image: \{\oc{\textbf{present / absent}}\}

\vspace{0.12em}
\textbf{\# Candidates (Existing Memories):}\\
\textit{[For each candidate concept $c_k \in \mathcal{C}_K$, $k = 1, \ldots, K$:]}

\textbf{\#\# Candidate:} CANDIDATE\_\{\oc{\textbf{$k$}}\}\\
- Portrait: [IMAGE]\\
- Static Memory: \{\oc{\textbf{Static Memory Items}}\}\\
- Recent Dynamic Memory: \{\oc{\textbf{Recent Dynamic Memory Items}}\}

\vspace{0.12em}
\textbf{\# Instructions}\\
Compare the entity in the Question against each Candidate.
\begin{itemize}[leftmargin=*, topsep=2pt, itemsep=0pt]
\item Check for mismatches in \textbf{Name}, \textbf{Species}, \textbf{Physical Appearance}, or \textbf{Core Attributes}.
\item If the Question describes an entity that is fundamentally different from a Candidate, they are NOT the same.
\item If the Question explicitly mentions a candidate's proper name, select that candidate unless species or stable identity details contradict it.
\item Transient details (e.g., accessories, location, mood, pose, or activity) may differ from memory and are not enough by themselves to make a new concept; compare stable identity details (e.g., name, species, breed, color, or markings).
\item If the tone of the Question is to introduce a new concept (e.g., a person, pet, or belonging), select ``NEW\_CONCEPT''.
\item Context might be similar, but specific details must match.
\item Use only the neutral candidate labels shown above. Do not infer identity from labels.
\item If NO candidate is a strong match, select ``NEW\_CONCEPT''.
\end{itemize}

\vspace{0.12em}
\textbf{\# Output Format}\\
Output ONLY one of the following:\\
1. ``EXISTING:CANDIDATE\_N'' (if it matches a candidate)\\
2. ``NEW\_CONCEPT''
\end{tcolorbox}

\end{figure*}

\begin{figure*}[!t]
\centering

\begin{tcolorbox}[
  colback=gray!10,
  colframe=black,
  boxrule=0.1mm,
  width=\textwidth,
  sharp corners,
  title=\textbf{Prompt $\mathcal{P}_f$ for Concept Fusion.}
]
\label{app:prompt-pf}

\textbf{Prompt:}
You are a concept similarity expert. Your task is to determine if two concepts represent the SAME entity that was incorrectly split (fissioned).

\vspace{0.12em}
\textbf{\# CONCEPT A}\\
\textbf{\#\# Portrait:} \{\oc{\textbf{Portrait Image or ``(No portrait available)''}}\}\\
\textbf{\#\# Static Memory (Permanent Facts):} \{\oc{\textbf{Static Memory Items}}\}\\
\textbf{\#\# Dynamic Memory (Recent Events):} \{\oc{\textbf{Dynamic Memory Items}}\}

\vspace{0.12em}
\textbf{\# CONCEPT B}\\
\textbf{\#\# Portrait:} \{\oc{\textbf{Portrait Image or ``(No portrait available)''}}\}\\
\textbf{\#\# Static Memory (Permanent Facts):} \{\oc{\textbf{Static Memory Items}}\}\\
\textbf{\#\# Dynamic Memory (Recent Events):} \{\oc{\textbf{Dynamic Memory Items}}\}

\vspace{0.12em}
\textbf{\# Instructions}\\
Compare the two concepts carefully to determine if they are the SAME entity.

\vspace{0.1em}
\textbf{CRITICAL RULES FOR ``SAME'' VERDICT:}
\begin{enumerate}[leftmargin=*, topsep=2pt, itemsep=0pt]
\item Conclude ``SAME'' only with clear identity evidence: the portraits show the same individual/object, or the memories share a distinctive name, model, or owner-specific identifier.
\item Generic category overlap is not enough: two cats or two similar-looking people can still be different entities.
\item If both concepts have specific names, brands, models, colors, breeds, or personal details that differ, conclude ``DIFFERENT''.
\item A generic reference (e.g., ``the cat'') matches a specific concept only when the visual evidence or distinctive memories strongly connect them.
\item Merge an unnamed fragment with few memories into the richer concept when its later details are compatible and no identity evidence contradicts it.
\end{enumerate}

\vspace{0.1em}
\textbf{GUIDELINES:}
\begin{itemize}[leftmargin=*, topsep=2pt, itemsep=0pt]
\item Assume memories accumulate over time.
\item Do not merge based on broad class, scene, pose, or activity alone.
\item Later memory fragments may omit the name.
\item If the evidence is weak or ambiguous, favor ``DIFFERENT''.
\end{itemize}

\vspace{0.12em}
\textbf{\# Output Format}\\
1. Brief reasoning (1--2 sentences).\\
2. Final verdict: ``VERDICT: SAME'' or ``VERDICT: DIFFERENT''.
\end{tcolorbox}

\end{figure*}

%% file: main.bib
@inproceedings{alaluf2024myvlm,
  title={Myvlm: Personalizing vlms for user-specific queries},
  author={Alaluf, Yuval and Richardson, Elad and Tulyakov, Sergey and Aberman, Kfir and Cohen-Or, Daniel},
  booktitle={European Conference on Computer Vision (ECCV)},
  pages={73--91},
  year={2024},
  organization={Springer}
}

@inproceedings{nguyen2024yollava,
	author = {Nguyen, Thao and Liu, Haotian and Li, Yuheng and Cai, Mu and Ojha, Utkarsh and Lee, Yong Jae},
	booktitle = {Advances in Neural Information Processing Systems (NeurIPS)},
	year = {2024},
	title = {Yo'{LLaVA}: Your {Personalized} {Language} and {Vision} {Assistant}.},
}

@inproceedings{hao2024rap,
	author = {Hao, Haoran and Han, Jiaming and Li, Changsheng and Li, Yu-Feng and Yue, Xiangyu},
	booktitle = {Proceedings of the Computer Vision and Pattern Recognition Conference (CVPR)},
	year = {2025},
	title = {{RAP}: Retrieval-{Augmented} {Personalization} for {Multimodal} {Large} {Language} {Models}},
}

@inproceedings{pi2025personalized,
	author = {Pi, Renjie and Zhang, Jianshu and Han, Tianyang and Zhang, Jipeng and Pan, Rui and Zhang, Tong},
	booktitle = {International Conference on Learning Representations (ICLR)},
	year = {2025},
	pages = {},
	organization = {},
	title = {Personalized {Visual} {Instruction} {Tuning}},
	volume = {},
}

@inproceedings{kim2025mmpb,
  title={MMPB: It’s Time for Multi-Modal Personalization},
  author={Kim, Jaeik and Kim, Woojin and Park, Woohyeon and Do, Jaeyoung},
year={2025},
  booktitle={Advances in Neural Information Processing Systems (NeurIPS)}
}

@inproceedings{hong2026tameing,
  author = {Hong, Rongpei and Lang, Jian and Wang, Yong and Zhong, Ting and Zhou, Fan},
  booktitle = {Proceedings of the ACM SIGKDD Conference on Knowledge Discovery and Data Mining (KDD)},
  year = {2026},
  title = {TAMEing Long Contexts in Personalization: Towards Training-Free and State-Aware MLLM Personalized Assistant},
}

@inproceedings{das2025training,
	author = {Das, Deepayan and Talon, Davide and Wang, Yiming and Mancini, Massimiliano and Ricci, Elisa},
	booktitle = {Proceedings of the IEEE/CVF International Conference on Computer Vision (ICCV)},
	year = {2025},
	title = {Training-Free Personalization via Retrieval and Reasoning on Fingerprints},
}

@article{yang2025qwen3,
  title={Qwen3 technical report},
  author={Yang, An and Li, Anfeng and Yang, Baosong and Zhang, Beichen and Hui, Binyuan and Zheng, Bo and Yu, Bowen and Gao, Chang and Huang, Chengen and Lv, Chenxu and others},
  journal={arXiv preprint arXiv:2505.09388},
  year={2025}
}

@misc{gpt52025openai,
  title        = {GPT-5 System Card},
  author       = {{OpenAI}},
  year         = {2025},
  howpublished = {https://openai.com/index/gpt-5-system-card},
  institution  = {OpenAI},
  month        = aug,
}

@article{seifi2025personalization,
	author = {Seifi, Soroush and Dorovatas, Vaggelis and Cassinelli, Matteo and Despinoy, Fabien and Reino, Daniel Olmeda and Aljundi, Rahaf},
	journal = {Transactions on Machine Learning Research (TMLR)},
	year = {2026},
	title = {Personalization {Toolkit}: Training {Free} {Personalization} of {Large} {Vision} {Language} {Models}},
}

@article{wang2025internvl,
	author = {Wang, Weiyun and Gao, Zhangwei and Gu, Lixin and Pu, Hengjun and Cui, Long and Wei, Xingguang and others},
	journal = {arXiv preprint arXiv:2508.18265},
	year = {2025},
	title = {{InternVL3.5}: Advancing {Open}-{Source} {Multimodal} {Models} in {Versatility}, {Reasoning}, and {Efficiency}},
}

@inproceedings{liu2024grounding,
	author = {Liu, Shilong and Zeng, Zhaoyang and Ren, Tianhe and Li, Feng and Zhang, Hao and Yang, Jie and Jiang, Qing and Li, Chunyuan and Yang, Jianwei and Su, Hang and Zhu, Jun and Zhang, Lei},
	booktitle = {European Conference on Computer Vision (ECCV)},
	year = {2024},
	pages = {38--55},
	organization = {},
	title = {Grounding {DINO}: Marrying {DINO} with {Grounded} {Pre}-training for {Open}-{Set} {Object} {Detection}.},
	volume = {},
}

@article{li2026qwen3vlembedding,
	author = {Li, Mingxin and Zhang, Yanzhao and Long, Dingkun and Chen, Keqin and Song, Sibo and Bai, Shuai and Yang, Zhibo and Xie, Pengjun and Yang, An and Liu, Dayiheng and Zhou, Jingren and Lin, Junyang},
	journal = {arXiv preprint arXiv:2601.04720},
	year = {2026},
	title = {{Qwen3-VL-Embedding} and {Qwen3-VL-Reranker}: A Unified Framework for State-of-the-Art Multimodal Retrieval and Ranking},
}

@article{liu2023visual,
  title={Visual Instruction Tuning},
  author={Liu, Haotian and Li, Chunyuan and Wu, Qingyang and Lee, Yong Jae},
  journal={Advances in Neural Information Processing Systems (NeurIPS)},
  volume={36},
  pages={34892--34916},
  year={2023}
}

@inproceedings{salemi2024lamp,
  title={Lamp: When large language models meet personalization},
  author={Salemi, Alireza and Mysore, Sheshera and Bendersky, Michael and Zamani, Hamed},
  booktitle={Annual Meeting of the Association for Computational Linguistics (ACL)},
  pages={7370--7392},
  year={2024}
}

@inproceedings{liu2025personaaware,
	author = {Liu, Dongshuo and Wu, Zhijing and Song, Dandan and Huang, Heyan},
	booktitle = {Annual Meeting of the Association for Computational Linguistics (ACL)},
	year = {2025},
	pages = {103--123},
	organization = {},
	title = {A {Persona}-{Aware} {LLM}-{Enhanced} {Framework} for {Multi}-{Session} {Personalized} {Dialogue} {Generation}},
	volume = {},
}

@inproceedings{magister2024personalization,
	author = {Magister, Lucie Charlotte and Metcalf, Katherine and Zhang, Yizhe and Hoeve, Maartje ter},
	booktitle = {Proceedings of the First Workshop on Large Language Model Memorization (L2M2)},
	year = {2025},
	pages = {61--77},
	organization = {Association for Computational Linguistics},
	title = {On the {Way} to {LLM} {Personalization}: Learning to {Remember} {User} {Conversations}},
}

@article{afzoon2024persobench,
	author = {Afzoon, Saleh and Naseem, Usman and Beheshti, Amin and Jamali, Zahra},
	journal = {arXiv preprint arXiv:2410.03198},
	year = {2024},
	title = {{PersoBench}: Benchmarking {Personalized} {Response} {Generation} in {Large} {Language} {Models}},
}

@inproceedings{jiang2025know,
	author = {Jiang, Bowen and Hao, Zhuoqun and Cho, Young-Min and Li, Bryan and Yuan, Yuan and Chen, Sihao and Ungar, Lyle and Taylor, C. J. and Roth, Dan},
	booktitle = {Conference on Language Modeling (COLM)},
	year = {2025},
	title = {Know {Me}, {Respond} to {Me}: Benchmarking {LLMs} for {Dynamic} {User} {Profiling} and {Personalized} {Responses} at {Scale}},
}

@inproceedings{an2025unictokens,
  title={{UniCTokens}: Boosting Personalized Understanding and Generation via Unified Concept Tokens},
  author={An, Ruichuan and Yang, Sihan and Zhang, Renrui and Shen, Zijun and Lu, Ming and Dai, Gaole and Liang, Hao and Guo, Ziyu and Yan, Shilin and Luo, Yulin and others},
  booktitle={Advances in Neural Information Processing Systems (NeurIPS)},
  year={2025}
}

@inproceedings{nguyen2025yo,
  title={Yo'Chameleon: Personalized Vision and Language Generation},
  author={Nguyen, Thao and Singh, Krishna Kumar and Shi, Jing and Bui, Trung and Lee, Yong Jae and Li, Yuheng},
  booktitle={Proceedings of the Computer Vision and Pattern Recognition Conference (CVPR)},
  pages={14438--14448},
  year={2025}
}

@inproceedings{zhang2018personalizing,
	author = {Zhang, Saizheng and Dinan, Emily and Urbanek, Jack and Szlam, Arthur and Kiela, Douwe and Weston, Jason},
	booktitle = {Annual Meeting of the Association for Computational Linguistics (ACL)},
	year = {2018},
	pages = {2204--2213},
	organization = {},
	title = {Personalizing {Dialogue} {Agents}: I have a dog, do you have pets too?},
	volume = {},
}

@inproceedings{jang2022call,
	author = {Jang, Yoonna and Lim, Jungwoo and Hur, Yuna and Oh, Dongsuk and Son, Suhyune and Lee, Yeonsoo and Shin, Dong-Hoon and Kim, Seungryong and Lim, Heuiseok},
	booktitle = {Proceedings of the AAAI Conference on Artificial Intelligence (AAAI)},
	year = {2022},
	pages = {10803--10812},
	organization = {},
	title = {Call for {Customized} {Conversation}: Customized {Conversation} {Grounding} {Persona} and {Knowledge}.},
	volume = {},
}

@inproceedings{huang2025learn,
	author = {Huang, Wenke and Liang, Jian and Shi, Zekun and Zhu, Didi and Wan, Guancheng and Li, He and Du, Bo and Tao, Dacheng and Ye, Mang},
	booktitle = {International Conference on Machine Learning (ICML)},
	year = {2025},
	pages = {},
	organization = {},
	title = {Learn from {Downstream} and {Be} {Yourself} in {Multimodal} {Large} {Language} {Models} {Fine}-{Tuning}},
	volume = {},
}

@inproceedings{lian2025affectgpt,
	author = {Lian, Zheng and Chen, Haoyu and Chen, Lan and Sun, Haiyang and Sun, Licai and Ren, Yong and Cheng, Zebang and Liu, Bin and Liu, Rui and Peng, Xiaojiang and Yi, Jiangyan and Tao, Jianhua},
	booktitle = {International Conference on Machine Learning (ICML)},
	year = {2025},
	pages = {},
	organization = {},
	title = {{AffectGPT}: A {New} {Dataset}, {Model}, and {Benchmark} for {Emotion} {Understanding} with {Multimodal} {Large} {Language} {Models}},
	volume = {},
}

@inproceedings{jung2025visual,
	author = {Jung, Mingi and Lee, Saehyung and Kim, Eunji and Yoon, Sungroh},
	booktitle = {International Conference on Machine Learning (ICML)},
	year = {2025},
	pages = {},
	organization = {},
	title = {Visual {Attention} {Never} {Fades}: Selective {Progressive} {Attention} {ReCalibration} for {Detailed} {Image} {Captioning} in {Multimodal} {Large} {Language} {Models}},
	volume = {},
}

@inproceedings{chang2025lvxattn,
	author = {Chang, Tzu-Tao and Venkataraman, Shivaram},
	booktitle = {International Conference on Machine Learning (ICML)},
	year = {2025},
	pages = {},
	organization = {},
	title = {{LV-XAttn}: Distributed {Cross}-{Attention} for {Long} {Visual} {Inputs} in {Multimodal} {Large} {Language} {Models}},
	volume = {},
}

@inproceedings{gu2024mllmguard,
	author = {Gu, Tianle and Zhou, Zeyang and Huang, Kexin and Liang, Dandan and Wang, Yixu and Zhao, Haiquan and Yao, Yuanqi and Qiao, Xingge and Wang, Keqing and Yang, Yujiu and Teng, Yan and Qiao, Yu and Wang, Yingchun},
	booktitle = {Advances in Neural Information Processing Systems (NeurIPS)},
	year = {2024},
	pages = {},
	organization = {},
	title = {MLLMGuard: A {Multi}-dimensional {Safety} {Evaluation} {Suite} for {Multimodal} {Large} {Language} {Models}.},
	volume = {},
}

@inproceedings{liu2024iibench,
	author = {Liu, Ziqiang and Fang, Feiteng and Feng, Xi and Du, Xeron and Zhang, Chenhao and Wang, Noah and Bai, Yuelin and Zhao, Qixuan and Fan, Liyang and Gan, Chengguang and Lin, Hongquan and Li, Jiaming and Ni, Yuansheng and Wu, Haihong and Narsupalli, Yaswanth and Zheng, Zhigang and Li, Chengming and Hu, Xiping and Xu, Ruifeng and Chen, Xiaojun and Yang, Min and Liu, Jiaheng and Liu, Ruibo and Huang, Wenhao and Zhang, Ge and Ni, Shiwen},
	booktitle = {Advances in Neural Information Processing Systems (NeurIPS)},
	year = {2024},
	pages = {},
	organization = {},
	title = {II-{Bench}: An {Image} {Implication} {Understanding} {Benchmark} for {Multimodal} {Large} {Language} {Models}.},
	volume = {},
}

@inproceedings{luo2024feast,
	author = {Luo, Gen and Zhou, Yiyi and Zhang, Yuxin and Zheng, Xiawu and Sun, Xiaoshuai and Ji, Rongrong},
	booktitle = {International Conference on Learning Representations (ICLR)},
	year = {2025},
	pages = {},
	organization = {},
	title = {Feast {Your} {Eyes}: Mixture-of-{Resolution} {Adaptation} for {Multimodal} {Large} {Language} {Models}},
	volume = {},
}

@inproceedings{kao2024bridging,
	author = {Kao, Chia-Hao and Chien, Cheng and Tseng, Yu-Jen and Chen, Yi-Hsin and Gnutti, Alessandro and Lo, Shao-Yuan and Peng, Wen-Hsiao and Leonardi, Riccardo},
	booktitle = {International Conference on Learning Representations (ICLR)},
	year = {2024},
	pages = {},
	publisher = {},
	title = {Bridging {Compressed} {Image} {Latents} and {Multimodal} {Large} {Language} {Models}},
	volume = {},
}

@inproceedings{liu2024reasoning,
  title={Reasoning multi-agent behavioral topology for interactive autonomous driving},
  author={Liu, Haochen and Chen, Li and Qiao, Yu and Lv, Chen and Li, Hongyang},
  booktitle={Advances in Neural Information Processing Systems (NeurIPS)},
  volume={37},
  pages={92605--92637},
  year={2024}
}

@inproceedings{wang20253d,
  title={3D Question Answering via only 2D Vision-Language Models},
  author={WANG, FENGYUN and Yu, Sicheng and Wu, Jiawei and Tang, Jinhui and Zhang, Hanwang and Sun, Qianru},
  booktitle={International Conference on Machine Learning (ICML)},
year={2025}
}

@inproceedings{zhu2025mmedpo,
  title={MMedPO: Aligning Medical Vision-Language Models with Clinical-Aware Multimodal Preference Optimization},
  author={Zhu, Kangyu and Xia, Peng and Li, Yun and Zhu, Hongtu and Wang, Sheng and Yao, Huaxiu},
  booktitle={International Conference on Machine Learning (ICML)},
year={2025}
}

@inproceedings{shen2024taskbench,
  title={Taskbench: Benchmarking large language models for task automation},
  author={Shen, Yongliang and Song, Kaitao and Tan, Xu and Zhang, Wenqi and Ren, Kan and Yuan, Siyu and Lu, Weiming and Li, Dongsheng and Zhuang, Yueting},
  booktitle={Advances in Neural Information Processing Systems (NeurIPS)},
  volume={37},
  pages={4540--4574},
  year={2024}
}

@article{chang2024survey,
  title={A survey on evaluation of large language models},
  author={Chang, Yupeng and Wang, Xu and Wang, Jindong and Wu, Yuan and Yang, Linyi and Zhu, Kaijie and Chen, Hao and Yi, Xiaoyuan and Wang, Cunxiang and Wang, Yidong and others},
  journal={ACM Transactions on Intelligent Systems and Technology (TIST)},
  volume={15},
  number={3},
  pages={1--45},
  year={2024},
  publisher={ACM New York, NY}
}

@inproceedings{zhao2024expel,
	author = {Zhao, Andrew and Huang, Daniel and Xu, Quentin and Lin, Matthieu and Liu, Yong-Jin and Huang, Gao},
	booktitle = {Proceedings of the AAAI Conference on Artificial Intelligence (AAAI)},
	year = {2024},
	pages = {19632--19642},
	organization = {},
	title = {ExpeL: LLM {Agents} {Are} {Experiential} {Learners}.},
	volume = {},
}

@article{zhang2025survey,
	author = {Zhang, Zeyu and Dai, Quanyu and Bo, Xiaohe and Ma, Chen and Li, Rui and Chen, Xu and Zhu, Jieming and Dong, Zhenhua and Wen, Ji-Rong},
	journal = {ACM Transactions on Information Systems},
	number = {6},
	year = {2025},
	pages = {1--47},
	publisher = {Association for Computing Machinery (ACM)},
	title = {A {Survey} on the {Memory} {Mechanism} of {Large} {Language} {Model}-based {Agents}},
	volume = {43},
}

@inproceedings{maharana2024evaluating,
	author = {Maharana, A. and Lee, Dong-Ho and Tulyakov, S. and Bansal, Mohit and Barbieri, Francesco and Fang, Yuwei},
	booktitle = {Annual Meeting of the Association for Computational Linguistics (ACL)},
	year = {2024},
	pages = {13851--13870},
	organization = {},
	title = {Evaluating {Very} {Long}-{Term} {Conversational} {Memory} of {LLM} {Agents}},
	volume = {},
}

@article{liu2023thinkinmemory,
	author = {Liu, Lei and Yang, Xiaoyan and Shen, Yue and Hu, Binbin and Zhang, Zhiqiang and Gu, Jinjie and Zhang, Guannan},
	journal = {arXiv preprint arXiv:2311.08719},
	year = {2023},
	title = {Think-in-{Memory}: Recalling and {Post}-thinking {Enable} {LLMs} with {Long}-{Term} {Memory}},
}

@inproceedings{zhong2024memorybank,
	author = {Zhong, Wanjun and Guo, Lianghong and Gao, Qiqi and Ye, He and Wang, Yanlin},
	booktitle = {Proceedings of the AAAI Conference on Artificial Intelligence (AAAI)},
	year = {2024},
	pages = {19724--19731},
	organization = {},
	title = {MemoryBank: Enhancing {Large} {Language} {Models} with {Long}-{Term} {Memory}.},
	volume = {},
}

@inproceedings{penha2024bridging,
	author = {Penha, Gustavo and Vardasbi, Ali and Palumbo, Enrico and Nadai, Marco De and Bouchard, Hugues},
	booktitle = {ACM Conference on Recommender Systems (RecSys)},
	year = {2024},
	pages = {340--349},
	organization = {ACM},
	title = {Bridging {Search} and {Recommendation} in {Generative} {Retrieval}: Does {One} {Task} {Help} the {Other}?},
	volume = {},
}

@inproceedings{xu2025memoryaugmented,
	author = {Xu, Mufan and Liang, Gewen and Chen, Kehai and Wang, Wei and Zhou, Xun and Yang, Muyun and Zhao, Tiejun and Zhang, Min},
	booktitle = {Findings of the Association for Computational Linguistics (ACL)},
	year = {2025},
	pages = {24068--24084},
	organization = {},
	title = {Memory-augmented {Query} {Reconstruction} for {LLM}-based {Knowledge} {Graph} {Reasoning}},
	volume = {},
}

@article{fu2025reason,
	author = {Fu, Yimin and Liu, Zhunga and Lyu, Jialin},
	journal = {IEEE Transactions on Pattern Analysis and Machine Intelligence (TPAMI)},
	number = {7},
	year = {2025},
	pages = {5586--5599},
	publisher = {IEEE},
	title = {Reason and {Discovery}: A {New} {Paradigm} for {Open} {Set} {Recognition}},
	volume = {47},
}

@inproceedings{liu2026exploratory,
	author = {Liu, Zeyuan and Kim, Jeonghye and Luo, Xufang and Li, Dongsheng and Yang, Yuqing},
	booktitle = {International Conference on Learning Representations (ICLR)},
	year = {2026},
	title = {Exploratory {Memory}-{Augmented} {LLM} {Agent} via {Hybrid} {On}- and {Off}-{Policy} {Optimization}},
}

@inproceedings{wang2026classifierinduced,
	author = {Wang, Yibo and Rui, Yong and Zhang, Min-Ling},
	booktitle = {Proceedings of the AAAI Conference on Artificial Intelligence (AAAI)},
	year = {2026},
	pages = {26643--26651},
	organization = {},
	title = {Classifier-induced {Reciprocal} {Points} for {Multi}-label {Open}-set {Recognition}},
	volume = {},
}

@article{scheirer2012toward,
  title={Toward open set recognition},
  author={Scheirer, Walter J and de Rezende Rocha, Anderson and Sapkota, Archana and Boult, Terrance E},
  journal={IEEE Transactions on Pattern Analysis and Machine Intelligence (TPAMI)},
  volume={35},
  number={7},
  pages={1757--1772},
  year={2012},
  publisher={IEEE}
}

@inproceedings{oza2019c2ae,
  title={C2ae: Class conditioned auto-encoder for open-set recognition},
  author={Oza, Poojan and Patel, Vishal M},
  booktitle={Proceedings of the Computer Vision and Pattern Recognition Conference (CVPR)},
  pages={2307--2316},
  year={2019}
}

@inproceedings{sun2020conditional,
  title={Conditional gaussian distribution learning for open set recognition},
  author={Sun, Xin and Yang, Zhenning and Zhang, Chi and Ling, Keck-Voon and Peng, Guohao},
  booktitle={Proceedings of the Computer Vision and Pattern Recognition Conference (CVPR)},
  pages={13480--13489},
  year={2020}
}

@inproceedings{guo2021conditional,
  title={Conditional variational capsule network for open set recognition},
  author={Guo, Yunrui and Camporese, Guglielmo and Yang, Wenjing and Sperduti, Alessandro and Ballan, Lamberto},
  booktitle={Proceedings of the IEEE/CVF International Conference on Computer Vision (ICCV)},
  pages={103--111},
  year={2021}
}

@inproceedings{zhou2021learning,
  title={Learning placeholders for open-set recognition},
  author={Zhou, Da-Wei and Ye, Han-Jia and Zhan, De-Chuan},
  booktitle={Proceedings of the Computer Vision and Pattern Recognition Conference (CVPR)},
  pages={4401--4410},
  year={2021}
}

@inproceedings{xu2023contrastive,
  title={Contrastive open set recognition},
  author={Xu, Baile and Shen, Furao and Zhao, Jian},
  booktitle={Proceedings of the AAAI Conference on Artificial Intelligence (AAAI)},
  volume={37},
  number={9},
  pages={10546--10556},
  year={2023}
}

@inproceedings{hang2024binary,
  title={Binary decomposition: A problem transformation perspective for open-set semi-supervised learning},
  author={Hang, Jun-Yi and Zhang, Min-Ling},
  booktitle={International Conference on Machine Learning (ICML)},
  year={2024}
}

@inproceedings{parkhi2012cats,
	author = {Parkhi, Omkar M. and Vedaldi, Andrea and Zisserman, Andrew and Jawahar, C. V.},
	booktitle = {Proceedings of the Computer Vision and Pattern Recognition Conference (CVPR)},
	year = {2012},
	pages = {3498--3505},
	organization = {},
	title = {Cats and dogs.},
	volume = {},
}

@inproceedings{lin2014microsoft,
	author = {Lin, Tsung-Yi and Maire, Michael and Belongie, Serge J. and Hays, James and Perona, Pietro and Ramanan, Deva and Doll{\' a}r, Piotr and Zitnick, C. Lawrence},
	booktitle = {European Conference on Computer Vision (ECCV)},
	year = {2014},
	pages = {740--755},
	organization = {},
	title = {Microsoft {COCO}: Common {Objects} in {Context}.},
	volume = {},
}

@misc{qwen3.5,
    title  = {{Qwen3.5}: Towards Native Multimodal Agents},
    author = {{Qwen Team}},
    month  = {February},
    year   = {2026},
    url    = {https://qwen.ai/blog?id=qwen3.5}
}

@inproceedings{nie2026personavlm,
  title={PersonaVLM: Long-Term Personalized Multimodal LLMs},
  author={Nie, Chang and Fu, Chaoyou and Zhang, Yifan and Yang, Haihua and Shan, Caifeng},
  booktitle={Proceedings of the Computer Vision and Pattern Recognition Conference (CVPR)},
  year={2026},
  url={http://arxiv.org/abs/2604.13074}
}

@inproceedings{an2026concept,
  title={{Concept-as-Tree}: A Controllable Synthetic Data Framework Makes Stronger Personalized {VLMs}},
  author={An, Ruichuan and Zeng, Kai and Lu, Ming and Yang, Sihan and Zhang, Renrui and Ji, Huitong and Liang, Hao and Zhang, Wentao},
  booktitle={European Conference on Computer Vision (ECCV)},
  year={2026}
}

@inproceedings{pham2025plvm,
  title={{PLVM}: A Tuning-Free Approach for Personalized Large Vision-Language Model},
  author={Pham, Chau and Phan, Hoang and Doermann, David and Tian, Yunjie},
  booktitle={Proceedings of the IEEE/CVF Conference on Computer Vision and Pattern Recognition Workshops (CVPRW)},
  pages={3671--3680},
  year={2025}
}

@article{an2024mcllava,
  title={{MC-LLaVA}: Multi-Concept Personalized Vision-Language Model},
  author={An, Ruichuan and Yang, Sihan and Zhang, Renrui and Lu, Ming and Jiang, Tianyi and Zeng, Kai and Luo, Yulin and Cao, Jiajun and Liang, Hao and Chen, Ying and She, Qi and Zhang, Shanghang and Zhang, Wentao},
  journal={arXiv preprint arXiv:2411.11706},
  year={2024}
}

@inproceedings{lang2026shallow,
  title={From shallow humor to metaphor: Towards label-free harmful meme detection via lmm agent self-improvement},
  author={Lang, Jian and Hong, Rongpei and Zhong, Ting and Chen, Leiting and Gao, Qiang and Zhou, Fan},
  booktitle={Proceedings of the 32nd ACM SIGKDD Conference on Knowledge Discovery and Data Mining V. 1},
  pages={588--599},
  year={2026}
}
